%% file: custom.tex
\pdfoutput=1

\documentclass[11pt]{article}

\usepackage[preprint]{neurips}

\usepackage{threeparttable}
\usepackage{times}
\usepackage{hyperref}
\usepackage{natbib}
\usepackage{latexsym}
\usepackage{caption}
\usepackage{subfigure}
\usepackage{siunitx}
\usepackage{physics}
\usepackage{amsmath}
\usepackage{tikz}
\usepackage{yhmath}
\usepackage{cancel}
\usepackage{color}
\usepackage{array}
\usepackage{multirow}
\usepackage{amssymb}
\usepackage{gensymb}
\usepackage{tabularx}
\usepackage{extarrows}
\usepackage{booktabs}
\usepackage{tabularx}
\usepackage{pgfplots}
\usetikzlibrary{fadings}
\usetikzlibrary{patterns}
\usetikzlibrary{shadows.blur}
\usetikzlibrary{shapes}
\usepackage{colortbl}
\usepackage{xcolor}
\usepackage{tipa}
\usepackage{fontawesome}
\usepackage{pifont} 
\definecolor{lgreen}{rgb}{0.937,0.992,0.929}
\definecolor{dgreen}{rgb}{0.470,0.650,0.369}
\definecolor{lblue}{rgb}{0.902,0.933,1.000}
\definecolor{llblue}{RGB}{235,242,247}
\definecolor{dblue}{RGB}{165,195,229}
\definecolor{lyellow}{RGB}{252,234,185}
\definecolor{dodgerblue}{rgb}{0.117,0.564,1.000}
\definecolor{orangered}{rgb}{1.000,0.270,0.000}
\definecolor{nasdaqup}{rgb}{0.000,0.654,0.356}
\usepgfplotslibrary{groupplots} 
\usepgfplotslibrary[groupplots]
\usetikzlibrary{pgfplots.groupplots}
\usetikzlibrary[pgfplots.groupplots]
\usetikzlibrary{fit}
\usetikzlibrary{backgrounds}

\usepackage{graphicx}
\usepackage{svg} 

\usepackage{tikz}
\usetikzlibrary{shapes.geometric, arrows}
\usepackage{pgfplots}
\pgfplotsset{compat=1.18}
\usepackage{tikz}
\usetikzlibrary{shapes.geometric, arrows, positioning}

\usepackage[T1]{fontenc}

\usepackage[utf8]{inputenc}

\usepackage{microtype}

\usepackage{inconsolata}

\usepackage{graphicx}
\usepackage{wrapfig}

\title{Soundwave\raisebox{-0.2em}{\includegraphics[height=1.2em]{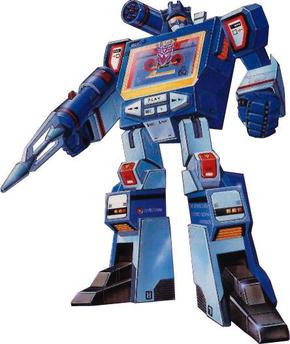}}: \textit{Less is More}  for Speech-Text Alignment in LLMs}

\author{ Yuhao Zhang, Zhiheng Liu, Fan Bu,  Ruiyu Zhang, Benyou Wang\thanks{Benyou is the corresponding author.}~, Haizhou Li\\
The Chinese University of Hong Kong, Shenzhen\\
\textit{yoohao.zhang@gmail.com, wangbenyou@cuhk.edu.cn} \\
}

\begin{document}
\maketitle
\begin{abstract}
Existing end-to-end speech large language models (LLMs) usually rely on large-scale annotated data for training, while data-efficient training has not been discussed in depth. We focus on two fundamental problems between speech and text: the representation space gap and sequence length inconsistency. We propose Soundwave, which utilizes an efficient training strategy and a novel architecture to address these issues. Results show that Soundwave outperforms the advanced Qwen2-Audio in speech translation and AIR-Bench speech tasks, using only one-fiftieth of the training data. Further analysis shows that Soundwave still retains its intelligence during conversation. The project is available at \href{https://github.com/FreedomIntelligence/Soundwave}{https://github.com/FreedomIntelligence/Soundwave}. 
\end{abstract}

\section{Introduction}

Large language models (LLMs) have profoundly transformed the paradigm of natural language processing (NLP) due to their remarkable abilities in understanding and reasoning \citep{achiam2023gpt,touvron2023llama}. Recently, multi-modal LLMs have also shown rapid development, with the success of GPT-4o highlighting the potential of speech-focused LLMs \citep{hurst2024gpt}. A fundamental requirement for achieving seamless communication with LLMs is their ability to accurately interpret speech—essentially enabling LLMs to ``hear''.

\begin{figure}[t]
    \centering
    \vspace{-10pt}
    \includegraphics[width=0.6\linewidth]{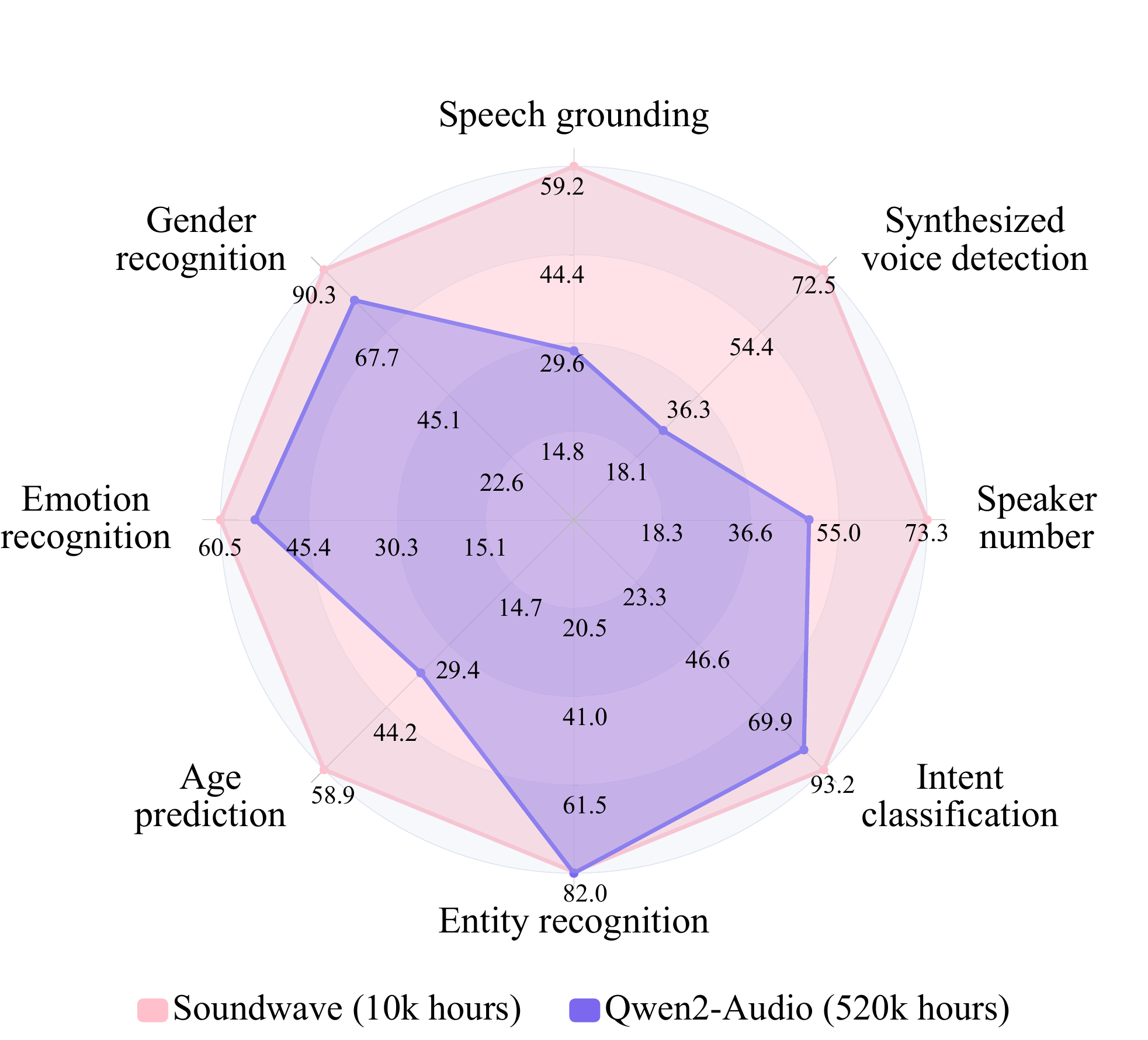}
    \caption{AIR-Bench speech foundation tasks.}
    \label{AIRbenchSpeech}
    \vspace{-0.4cm}
\end{figure}

However, most speech-based LLMs rely on massive labeled datasets and substantial computational resources to enable speech perception \citep{defossez2024moshi, chen2025minmo}. For example, the Qwen2-Audio \citep{qwen2audio} model requires approximately 500,000 hours of data to achieve cross-modal functionality, while 1,000 hours needed by advanced automatic speech recognition models to achieve comparable results \citep{gulati2020conformer}. This discrepancy underscores the need for more efficient methods to develop speech-capable LLMs.

We first identify two fundamental challenges to achieve alignment between speech and text \citep{zhang2023rethinking}: (1) the \textbf{representation space gap} and (2) \textbf{sequence length inconsistency}. The former challenge arises from the two modalities being developed independently, while the second challenge stems from the difference in modeling units—speech is typically represented at the frame level, whereas LLMs operate at the sub-word level. We then decouple the process of building speech LLMs to achieve more efficient training. Specifically, we propose a two-stage training framework designed to efficiently overcome these challenges. The first stage focuses on resolving the representation space gap, while the second stage aims to reduce the sequence length of speech.

Furthermore, to efficiently bridge the gap between speech and text, the quality of alignment data plays a crucial role. To address this, we collect high-quality speech recognition data and manually annotate audio labels to support the first stage. For the second stage, we analyze the proportion of text data to ensure a smooth learning process. During the supervised fine-tuning (SFT) stage, we employ temperature sampling to balance the variety of tasks effectively.

We conduct experiments on various speech-to-text tasks and several sound-related tasks. We also compare our model with the strong system Qwen2-Audio on both closed- and open-ended tasks. Our average results achieve state-of-the-art performance on the AIR-Bench \citep{airbench} speech foundation task as shown in Fig.~\ref{AIRbenchSpeech}. It also shows comparable results on audio-related tasks. Furthermore, our method exhibits significant performance in zero-shot speech translation, demonstrating that Soundwave unlocks the full potential of LLMs. Soundwave delivers better performance with less training data, lower training costs, and fewer speech sequences.

Our main \textbf{contributions} are as follows. 1) We propose an efficient training framework that utilizes only ten thousand hours of training data to achieve state-of-the-art speech understanding performance. 2) We introduce dynamic multi-task learning in the post-training stage to enhance speech modeling and leverage the benefits of text data. 3) We annotate a sound classification dataset to support the alignment between sound and text, and provide high-quality instruction data based on the thinking process for speech LLMs.

\section{Methodology}

\begin{wraptable}[16]{r}{0.47\textwidth}
    \centering
    \footnotesize
    \vspace{-0.4cm}
    \caption{The parameters of different modules. The {\color{orange!60}orange} represents the number of training parameters.}
    \label{Parameters}
    \resizebox{0.48\textwidth}{!}{
    \setlength{\tabcolsep}{0.5mm}{
    \begin{tabular}{lrcl}
        \toprule
        \multirow{2}{*}{\textbf{Modules}}&\multirow{2}{*}{\textbf{\#Param.}}&\textbf{Training}&\multirow{2}{*}{\textbf{Details}}  \\
        &&\textbf{stage}& \\
        \midrule 
        Audio encoder&$\sim$635M&-&Whisper Large V3  \\
        \midrule
        Alignment&\multirow{2}{*}{\color{orange!60}$\sim$144M}&\multirow{2}{*}{I\&II}& One projection and \\
         adapter&&&Transformer layer\\
        \midrule
        Shrinking &\multirow{2}{*}{\color{orange!60}$\sim$67M}&\multirow{2}{*}{II}& One cross-attention\\
        adapter&&&and layer-norm \\
        \midrule
        LLMs& $\sim$8B&-&Llama3.1 \\
        \midrule
        LLM adapter&{\color{orange!60}$\sim$55M} &II\&III& LoRA \\
        \midrule
        Total&$\sim$9B&\\
        \bottomrule
    \end{tabular}
    }
    }

\end{wraptable}

\subsection{Overall Design}

The training process consists of three stages, as shown in Fig.~\ref{architecture}. Stage I aims to align the representation between speech and text, addressing the \textit{representation space gap} problem. Stage II primarily shrinks the speech sequence and mitigates the \textit{sequence length inconsistency}. The supervised fine-tuning (SFT) stage \citep{wei2021finetuned} enables the speech LLMs to generalize across diverse tasks.

The input to the model consists of speech FBank features, which are then processed by the pretrained audio encoder. To efficiently align the representation with that of LLMs, we use the audio encoder that produces semantic features (e.g., Whisper \citep{whisper} or Seamless \citep{seamless2025joint}), rather than vector quantization features \citep{defossez2022high} or self-supervised features \citep{hubert}. We implement an alignment adapter and a shrinking adapter to bridge the gap between speech and text. Additionally, LoRA \citep{LoRA} is used to enable efficient fine-tuning. An overview of the modules is provided in Tab.~\ref{Parameters}.

\subsection{Stage I: Alignment}
\label{sec:alignment}

We use the auxiliary CTC loss to improve training efficiency, as it can achieve alignment without the involvement of LLMs. Additionally, we use high-quality data to speed up the convergence rate.
\subsubsection{Auxiliary CTC loss}

The audio encoder and LLMs have a gap in their representation spaces due to separate pre-training. One direct approach is to use ASR tasks for alignment. We design an adapter and utilize CTC loss \citep{graves2006connectionist} to achieve efficient cross-modal training. Specifically, the adapter consists of a linear layer followed by a Transformer layer \citep{vaswani2017attention}. The linear layer transforms the audio sequence $A\in \mathbb{R}^{l\times h_{a}}$ where $l$ is the length of the speech sequence and $h_{a}$ is the hidden size of the audio encoder. We concatenate adjacent features and adjust the dimensionality to match that of the LLMs, resulting in $A'\in \mathbb{R}^{l/2\times h_{llm}}$ where $h_{llm}$ is the hidden size of the LLMs. A Transformer layer then converts the features into the representation space of LLMs. Finally, we use CTC loss to train the adapter, aligning the shared space of the LLMs.

\input{arch}

\subsubsection{High-quality Alignment Data}

We believe that \textit{improving data quality is crucial to training efficiency for alignment}. We apply data strategies for two types of data (ASR and sound data), as outlined below. The adapter is trained without the LLMs at this stage, thus the alignment training is fast. Our later experiments in Sec. \ref{sec:data_quality} confirm the benefits to training efficiency.

\paragraph{Verified ASR Data}
At this training stage, we use transcriptions from ASR data as the target, which we found to be crucial for improving convergence ratio. The selected high-quality data is all verified by advanced ASR model \citep{whisper} with a Word Error Rate (WER) lower than 10\%. 

\paragraph{Standardized Sound Data}
Another challenge is processing sound due to the inherent background noise and the diversity of labeling information. To address this, we annotate about 8k pieces of sound category data. We further select clear 20k sound samples, then unify label format and audio length. 

\subsection{Stage II: Shrinking}

After aligning the data representation, we focus on reducing the length of the speech as detailed in Sec. \ref{sec:shrinking}. Additionally, at this stage, we include various types of foundational audio tasks to better generalize downstream tasks. This introduces a data mixture problem, which is solved by a dynamic data mixture strategy (see Sec. \ref{sec:mixture}).

\subsubsection{Dynamic Shrinking}
\label{sec:shrinking}
\input{shrinking}

There are two essential aspects to shrinking the audio sequence: \textit{final length determination} and \textit{lossless information retention}.

\paragraph{Final Length Determination}
For the first aspect, we utilize the probability from CTC. The CTC predicts the corresponding word for each position. We then remove duplicate predictions from adjacent positions to obtain the final sequence. Since the sequence has been aligned to text in Stage I, the decoded result can indicate the final length.

\paragraph{Lossless Information Retention}
For the second aspect, we select the content based on the CTC output as the query, and then use attention mechanisms to gather related information, such as tone and pitch, in order to prevent information loss. 

Assume the speech features $x$ have been aligned to the representation space of LLMs, then, we select the features based on the CTC probability to compress the sequence $x'$.   
\begin{equation} 
    x_{\mathrm{out}} = \textrm{norm}\left(x' + \textrm{cross\_attn}(x', x, x)\right)
\end{equation}
where norm is the layer norm operation. $x_{\mathrm{out}}$ is the final output of the shrinking adapter. $x'$ can be viewed as the content feature, while the gathered information, calculated by cross-attention, serves as auxiliary data for the selected features. The whole processing is shown in Fig.~\ref{lbm_method}.
\begin{table}[h]
    \centering
    \small
    \caption{The overview of tasks in shrinking stage. The data scales of these data are highly imbalanced.}
    \label{Multi_task_data}
    \begin{tabular}{lllr}
        \toprule
        \textbf{Task}&\textbf{Input}&\textbf{Output}&\textbf{Size (k)}  \\
        \midrule 
        QA&Text question&Text answer&78 \\
        \midrule 
        ASR&Speech&Transcription&3,012 \\ \midrule 
        ST&Speech&Translation&460 \\ \midrule 
        \multirow{2}{*}{Sound}&Mixed speech &Transcription &\multirow{2}{*}{25}\\
        &and sound & and sound type&\\
        \bottomrule
    \end{tabular}

\end{table}

\input{data_info}

\subsubsection{Dynamic Data Mixture} 
\label{sec:mixture}

We select both audio data (involving three basic audio tasks) and text data to enable LLMs to generalize to downstream speech understanding. Training with mixed data may be biased by dominant tasks due to data imbalance (see Table~\ref{Multi_task_data}), and existing work has adopted curriculum learning~\citep{speechverse, SALMONN}, though it requires considerable prior knowledge for proper design.

Inspired by temperature-based data sampling, which has previously been used to address multilingual data imbalance~\citep{arivazhagan2019massively}, we propose a dynamic data mixture guided by sampling temperature. Specifically, the sample rate for each task $k$ is as follows:
\begin{equation} 
    p_{k} = \left( \frac{ \vert \mathcal{D}_{k} \vert }{\sum_i \vert \mathcal{D}_{i} \vert } \right)^{\frac{1}{T}}
\end{equation}
where the $\vert \mathcal{D}_{k} \vert$ denotes the data size of task $k$ and $T$ denotes the temperature. 
$T$ is initially set to 1 and gradually increases. This causes the training to start with a \textit{sample-level uniform distribution} and gradually shift to a \textit{task-level uniform distribution}. Training at the former stage might be dominated by rich-source tasks, while at the latter stage, training might be more balanced among tasks, potentially alleviating the over-fitting issue.

Additionally,  \citet{chen2024allava} shows that text-related tasks aid instruction following for multi-modal LLMs. We also introduce the text task to ensure a smoother cross-modal process. We incorporate the Wizard SFT dataset \citep{WizardLM} to help speech LLMs retain their understanding capabilities, thereby enhancing their ability to follow instructions for speech tasks.

\subsection{Stage III: Supervised Fine-tuning}

At this stage, we only fine-tune the parameters of LoRA, as speech and text are already aligned. Our goal is to enable the speech LLMs to handle more complex tasks and respond directly based on the speaker's speech. Thus, we use both text-based and speech-based instructions during SFT.

\section{Data Engineering}
\label{sec:data_augmentation}
We introduce the data details for the three stages,   respectively, see the summary in Tab.~\ref{datasets}. The data shown in the table has been cleaned and filtered, and the details of strategies can be found in App.~\ref{app:data_process}. We sample some speech from several dataset to control the quality and training cost. 

\subsection{Data During Stage I and II}

\paragraph{ASR Data} We choose high-quality datasets and filter the data with a WER of less than 10\%, as tested by Whisper medium. We apply SpecAugment \citep{park2019specaugment} to enhance the robustness of the model towards speech. To help  LLMs understand the conversation and the number of speakers, we splice speech from different speakers.  We denote the output format as `{\tt The first speaker says ... The second speaker says ...}'.

\paragraph{Sound Data}
The sound data is often too short and may be viewed as noise, which causes the model to fail in perceiving it. To address this problem, we embedded environmental sounds into the audio to construct the data. For example: `But there was a passenger dropped off for you, a little girl. \textit{<throat\_clearing>} It's a boy I've come for.' The special token is added to the conversation, and the model needs to transcribe both the sound and speech simultaneously. 
This method of learning both speech and audio also makes training efficient.

Another problem is that the sound is always a mix of multiple sound categories and exhibits constant repetition. This causes timing and label count issues, which increases the difficulty of learning alignment. Therefore, we standardize the audio to a duration of three seconds based on energy, and manually select and label the data. The detailed process of labeling can be found in our Appendix~\ref{sec:sound_process}.

\begin{table*}[h]
    \centering
    \small
    \caption{Examples of three QA formats for building SFT data in the speech emotion task.}
    \label{Text_instruction}
    \resizebox{1.\textwidth}{!}{
    \setlength{\tabcolsep}{2pt} 
    \begin{tabular}{lll}
        \toprule
        &\textbf{Instruction}&\textbf{Answer} \\
        \midrule 
        1&What feeling is the speaker trying to convey?&Neutral. \\ 
         \multirow{2}{*}{2}&What is the emotional theme of the speaker's message?& \multirow{2}{*}{The one answer you should pick is D. happiness.}\\
        &Pick one answer from: A. neutral B. fear C. surprise D. happiness. &\\
        3&What emotion is the speaker hinting at in their speech?&The speaker is hinting at surprise in their speech. \\
        \bottomrule
    \end{tabular}
    }
\end{table*} 
\subsection{Instruction Data During Stage III}

\paragraph{Text Instructions} The text-based instruction is designed to understand and analyze speech. We have created three types of QA formats, as shown in Tab.~\ref{Text_instruction}. The first requires the model to directly answer, which is the most difficult. The second provides detailed choices, and the last requires the model to output the answer in a natural format.

\begin{wrapfigure}[14]{r}{0.47\textwidth}
    \centering
    \includegraphics[width=\linewidth]
    {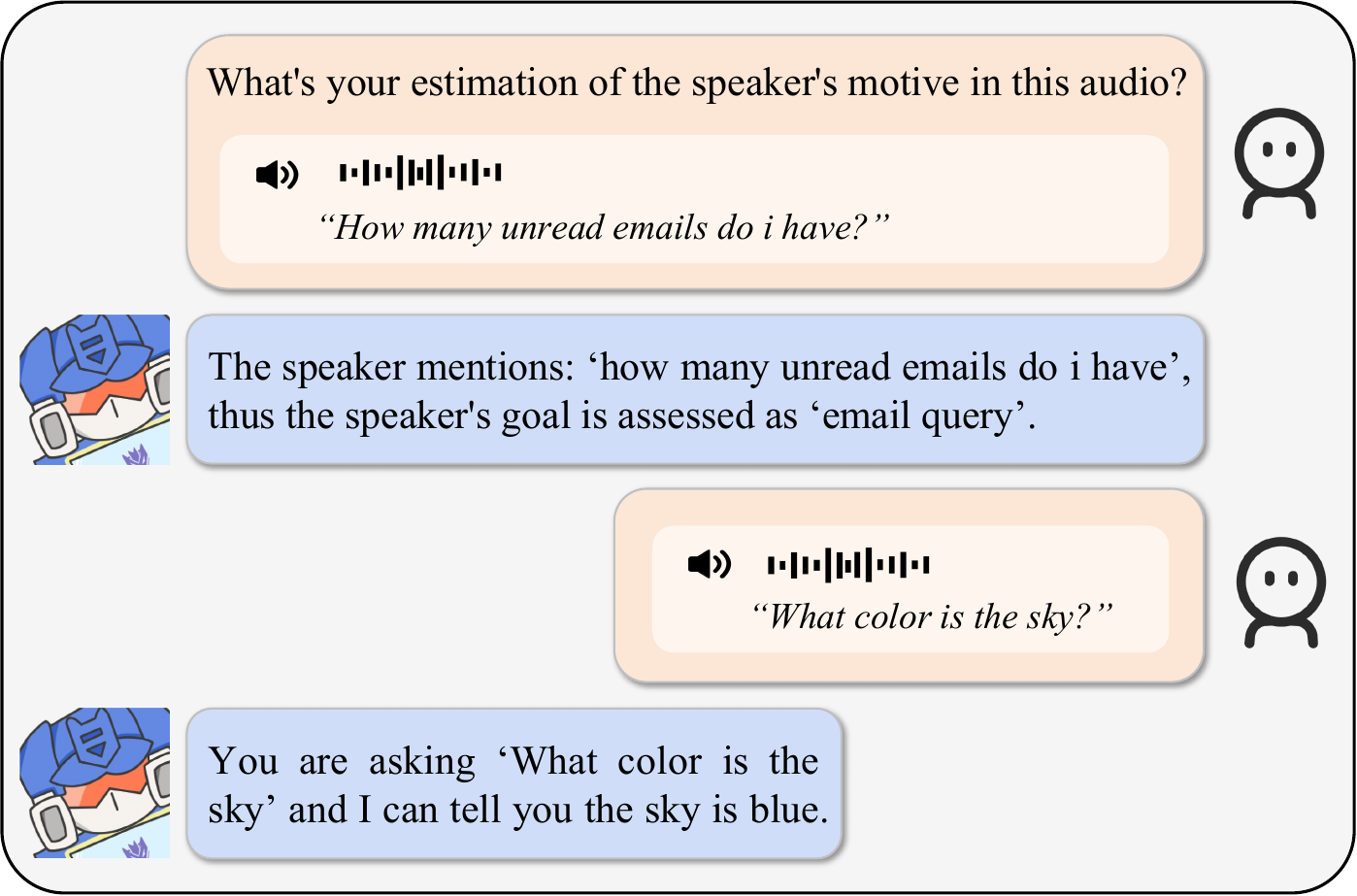}
    \caption{Adding thought processes to address complicated problems and speech instructions.}
    \label{COT}
    \vspace{-0.5cm}
\end{wrapfigure}

\paragraph{Speech Instructions} 
If speech LLMs are to communicate directly with humans, it is essential for them to follow speech instructions. Once the speech is well-aligned, we can achieve this by using text-based dialogue data and synthesizing text into speech. We use AnyInstruct speech subset \citep{zhan2024anygpt}, which is built using this approach.
 
\paragraph{Chain of Thought} To reduce the complexity of some challenging tasks, we built the dataset to enable the model to predict in a manner similar to Chain of Thought (CoT). For example, in the Intent Classification task, we first had the model output the speech transcription before identifying the intent, as shown in Fig.~\ref{COT}. For the speech instruction, the model initially predicts the transcription and then responds to the question. The reasoning time is slightly increased, but the model can address complicated tasks with limited training data.

\section{Experiments}

\subsection{Settings}
\label{sec:Experiments Setting}
\paragraph{Training}The audio encoder is Whisper Large V3 \citep{whisper}, and the foundation model is Llama-3.1-8B-Instruct \citep{dubey2024llama}. The alignment adapter is a projection where the output size is 4096. We apply LoRA to the Attention module, where rank and $\alpha$ are set to 64 and 16, respectively. Both alignment and shrinking stages consist of 6,000 steps, with the SFT stage set to around 4,000 steps. The sample temperature $T$ at Stage II starts at 1 and increases by 5 per training epoch. The experiments are conducted on 32 A800 GPUs for training on 10k hours of data. The training time for the two stages is approximately four days, and the SFT requires an additional day. App.~\ref{training_setting} shows more details about training settings.

\paragraph{Evaluation} 
We evaluate Soundwave on several basic tasks and the open-ended AIR-Bench. We also remove repeated samples (see App.~\ref{app:data_leak}) before training to avoid data leakage. We primarily compare Soundwave with Qwen2-Audio, an advanced model for various audio processing tasks.

\begin{table*}[t]
    \centering
    \small
    \caption{Performance on foundation tasks, including ASR, speech translation (ST), Speech Emotion Recognition (SER), Vocal Sound Classification (VSC). ST is evaluated by ScareBLEU \citep{post-2018-call}. $*$ denotes the zero-shot task. }
    \label{Foundation_task}
    \resizebox{1.\textwidth}{!}{
    \setlength{\tabcolsep}{2pt} 
    \begin{tabular}{lcccc}
        \toprule
        \textbf{Task}&\textbf{Dateset}&\textbf{Model}& \textbf{Metric}& \textbf{Performance}\\
        \midrule 
        \multirow{5}{*}{ASR}&\multirow{5}{*}{Librispeech (Test-clean|Test-other)}& SALMONN~\citep{SALMONN}&\multirow{5}{*}{WER $\downarrow$} &2.1 | 4.9\\ 
        && SpeechVerse~\citep{speechverse} & &2.5 | 4.7\\
        &&WavLLM~\citep{wavllm}& &2.0 | 4.8\\
        && Qwen2-Audio~\citep{qwen2audio}& &\textbf{1.6} | \textbf{3.6}\\
        && Soundwave &&2.1 | 5.0 \\ \cline{2-5}
        \hline
        \multirow{7}{*}{ST}&\multirow{4}{*}{CoVoST2 En-De}&BLSP~\citep{wang2023blsp} &\multirow{4}{*}{BLEU $\uparrow$}&14.1\\
        &&SALMONN~\citep{SALMONN}&&18.6\\
        &&Qwen2-Audio~\citep{qwen2audio}&&29.9\\
        &&Soundwave &&\textbf{30.6}\\ \cline{2-5}
        &\multirow{2}{*}{MuST-C$^*$ (En-Nl|En-It|En-Ro|En-Es)}&Qwen2-Audio ~\citep{qwen2audio}&\multirow{2}{*}{BLEU $\uparrow$}&20.7 | 19.5 | 11.8 | 22.1\\
        &&Soundwave&&\textbf{27.0} | \textbf{22.2} | \textbf{16.9} | \textbf{26.7}\\
        \hline
        \multirow{2}{*}{SER}&\multirow{2}{*}{Meld}& Qwen2-Audio~\citep{qwen2audio}&\multirow{2}{*}{ACC $\uparrow$}&0.553\\
        && Soundwave&&\textbf{0.635}\\
        \hline
        \multirow{3}{*}{VSC}&\multirow{3}{*}{VocalSound}& Pengi \citep{Pengi}&\multirow{3}{*}{ACC $\uparrow$}&0.604\\
        && Qwen2-Audio~\citep{qwen2audio}&&\textbf{0.939}\\
        && Soundwave&&0.905\\
        \bottomrule
    \end{tabular}
    }
    \vspace{-0.3cm}
\end{table*}
\subsection{Results}
\label{sec:Experiments Results}
\paragraph{Basic Audio Tasks}

We show the results on foundational audio tasks in Tab.~\ref{Foundation_task}. We find that our model demonstrates a significant advantage on the ST and SER tasks, which heavily rely on the understanding ability of speech LLMs. We also observe that our model shows strong performance on zero-shot tasks, such as translation tasks in other languages. On the other hand, our model still underperforms the SOTA model on the ASR task, indicating that massive training data is essential for ASR. We only used about 244 hours of sound data, which is dozens of times less than the SOTA, thus there is still a gap on the VSC task.

\begin{table*}[h]
    \centering
    \footnotesize
    \vspace{-5pt}
    \caption{Performance on the AIR-Bench speech foundation tasks.}
    \label{Airbench_speech}
    \resizebox{1.\textwidth}{!}{
    \setlength{\tabcolsep}{2pt}{
    \begin{tabular}{lcccccccccc}
        \toprule
        \multirow{2}{*}{\textbf{Task}}&\multirow{2}{*}{\textbf{Soundwave}}& \multirow{2}{*}{ \fontsize{8}{10}\selectfont \textbf{Qwen2-Audio}}&{\fontsize{8}{10}\selectfont \textbf{Qwen-Audio}}&\multirow{2}{*}{\textbf{SALMONN}}&\multirow{2}{*}{\textbf{NExT-GPT}}&\multirow{2}{*}{\textbf{PandaGPT}}&\textbf{Whisper} \\
        &&& {\fontsize{8}{10}\selectfont \textbf{Turbo}}&&&&\textbf{+GPT-4}\\
        \midrule 
        Speech Grounding&\textbf{59.2}&28.3&45.4&25.3& 25.4&23.0&35.0    \\ 
        Language Identification&{89.6}&{93.3}&{95.9}&{28.1}  &	{23.7}&	{34.6}	&{\textbf{96.8}}    \\
        Gender Recognition&{\textbf{90.3}}&{79.3}&{82.5}&{35.5} 	&{57.0}	&{66.5}	&{21.9}    \\
        Emotion Recognition& \textbf{60.5}&54.6&60.0&29.9	&25.7	&26.0	&59.5 \\ 
        Age Prediction&{58.9}&{36.1}&{58.8}&{48.7} 	&\textbf{62.4}	&{42.5}	&{41.1}        \\
        Entity Recognition&{81.7}&\textbf{82.0}&{48.1}&{51.7}     &{26.1}	&{34.0}	&{69.8}   \\
        Intent Classification&\textbf{93.2}&85.8&56.4&36.7	&25.6	&28.5	&87.7\\ 
        Speaker num. Verification& \textbf{73.3}&{48.8}&{54.3}&{34.3}  	&{25.4}	&{43.2}	&{30.0}    \\
        Synthesized Detection&\textbf{72.5}&{25.9}&{69.3}&{50.0}  	&{30.8}	&{53.1}	&{40.5}  \\
        Average&\textbf{75.5}&59.3&63.4&37.8&33.6&39.0&53.6\\ 
        \bottomrule
    \end{tabular}
    }
    }
\end{table*}
\paragraph{AIR-Bench}

We compare our model on AIR-Bench across speech foundation, sound foundation, and speech chat tasks. As shown in Tab.~\ref{Airbench_speech}, our model demonstrates SOTA performance on average speech foundation tasks with only about 10k of training data. Specifically, we outperform the best of previous speech LLMs on six sub-tasks. Since 98.61\% of the training data consists of English speech, our model performs worse on the language identification task. This highlights that the proportion of different languages remains important.

Results of the sound foundation task are shown in Tab.~\ref{Airbench_audio}. Although only around 244 hours of data were used, our model is still superior to other models, except Qwen2-Audio, which is trained with 10k hours. Moreover, our single-encoder architecture performs better than the two-encoder model \citep{SALMONN}, indicating that fewer encoders can process both speech and sound simultaneously. Our model also performs well in AIR-Bench speech chat task, ranking second only to Qwen2-Audio among open source models.

\begin{table*}[t]
    \centering
    \small
    \caption{Performance on the AIR-Bench sound foundation and chat tasks.}
    \label{Airbench_audio}
    \resizebox{1.\textwidth}{!}{
    \setlength{\tabcolsep}{2pt}{
    \begin{tabular}{lcccccccccc}
        \toprule
        \multirow{2}{*}{\textbf{Task}}&\multicolumn{1}{c}{\multirow{2}{*}{\textbf{Soundwave}}}& \multicolumn{1}{c}{\multirow{2}{*}{\textbf{Qwen2-Audio}}}&\multicolumn{1}{c}{\textbf{Qwen-Audio}}&\multicolumn{1}{c}{\multirow{2}{*}{\textbf{SALMONN}}}&\multicolumn{1}{c}{\multirow{2}{*}{\textbf{NExT-GPT}}}&\multicolumn{1}{c}{\multirow{2}{*}{\textbf{PandaGPT}}}&\textbf{Gemini} \\
        &&& \multicolumn{1}{c}{\textbf{Turbo}}&&&&\textbf{(1.5-pro)}\\
        \midrule 
        Sound (average)&62.10 &\textbf{65.10}&60.95&32.95&32.15&43.58&-    \\ 
        Speech Chat&\ \ 6.51&\ \ \textbf{7.18}&\ \ 7.04&\ \ 6.16&\ \ 3.86&\ \ 3.58&6.97    \\ 
        \bottomrule
    \end{tabular}
    }
    }
    \vspace{-0.2cm}
\end{table*}

\section{Analysis}
\label{Analysis}
Considering that analysis based on full data requires massive training cost, we analyze our method based on Librispeech data. The experiments are trained on 8 A800 GPUs with 4,000 steps. We use \textbf{Adapter $(\times n)$} to denote that the adapter architecture is the same as Qwen2-Audio, where $n$ is the down-sampling rate.

\input{model_compare_loss}
\subsection{Convergence Rate}

In Fig.~\ref{convergence_speed}, we compare the convergence rate with and without the first alignment stage, and the projection adapter architecture. Soundwave sees a high convergence rate, with the loss rapidly decreasing within the first hundred steps. In contrast, the training process of the other model is much slower without the alignment stage. Furthermore, Soundwave performs worse than other models without stage one, because the shrinking adapter relies on the CTC prediction.

\input{compare_sim_and_train_speed}
\subsection{Effect of Alignment}

We randomly sampled 200 items from the Librispeech test clean set and then extracted text and speech representations. The similarity of speech and text after average pooling is compared, as shown in Fig.~\ref{denosieperformance}. We found that the representation of Soundwave with the alignment adapter is significantly higher than that of other methods. In addition, we further compare the average training speed under the same batch conditions. The training speed in the alignment stage is nearly three times faster than that of other methods. Whether due to the alignment effect or the training method, the alignment adapter shows obvious advantages.

\subsection{Effect of Shrinking}

We compare the performance and compression ratios of different strategies on ASR tasks. We found that our approach compresses significantly based on text length. Our method maintains stable performance with 2.5\% compression ratios. However, the compression method leads to performance degradation on other test tasks without the aid of auxiliary information. This demonstrates that auxiliary information can compensate for missing features, allowing the LLMs to receive complete information.

\begin{table}[h]
    \centering
    \small
    \caption{Comparison of different shrinking methods on the Librispeech ASR dataset.}
    \label{tab:diff_shrinking_method}
    \begin{tabular}{lcccr}
        \toprule
        \textbf{Method}&\textbf{Test clean}&\textbf{Test other}&\textbf{TTFT (ms)}&\textbf{Compression ratio}\\ 
        \midrule 
        Shrinking adapter&3.1&6.6&72&2.5\% \\
         \ \ w/o auxiliary info.&3.1&7.1&72&2.5\% \\
        Adapter ($\times3$)&3.8&6.5&95&33.3\%\\
        Adapter ($\times4$)&4.3&7.8&85&25.0\%\\
        
        \bottomrule
    \end{tabular}

\end{table} 
We exhibit the inference speed in Tab.~\ref{tab:diff_shrinking_method}, using Time To First Token (TTFT) as the metric. Our method shows a speed-up of about 15\% and 25\% compared to Adapter ($\times3$) and Adapter ($\times4$) methods, respectively. This demonstrates that our method uses fewer tokens while achieving greater inference speed-up. We found that the shrinking adapter does not incur significant computational cost, proving it is both lightweight and effective.

\input{data_quality_loss_compare}

\subsection{Data Quality}
\label{sec:data_quality}

The training loss for Stage I, with and without cleaning the speech and sound data, is compared in Fig.~\ref{data_quality_convergence_speed}. When uncleaned speech is used, the training process becomes unstable. Additionally, if the sound data is not properly processed, it significantly worsens the overall training. Given that the alignment stage only trains a few parameters to align the two pretrained large models, the quality of the training data is crucial.

\input{model_compare_air_speech}
\subsection{Data Scaling}

We compare the performance from 1k to 10k hours of data, and the results are shown in Fig.~\ref{scaling_compare}. Our model, using only 1k hours of training data, achieves performance comparable to previous speech LLMs. Note that we use only the ASR task as the SFT data, yet our model demonstrates decent capability in instruction following. This demonstrates that the speech representation is well aligned with the text representation. When we further scale up the training data, all tasks show consistent improvements.

\subsection{Knowledge-Based QA}

\begin{figure}[h]
    \centering
    \includegraphics[width=0.6\textwidth]{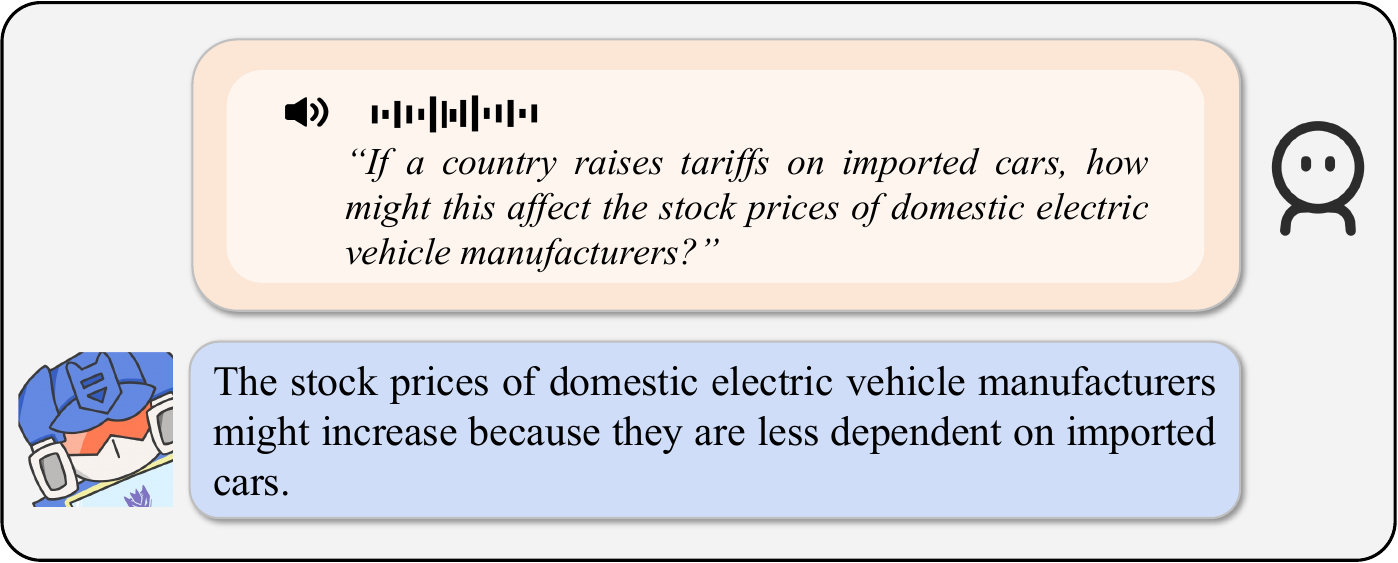}
    \caption{A case of answering the spoken question.}
    \label{pic: case}
    \vspace{-0.3cm}
\end{figure}

We present a case of using the speech instruction to ask complex questions in Fig.~\ref{pic: case}. We find that Soundwave inherits the rich knowledge of LLMs during the conversation. For more examples of performance in physics, chemistry, finance, mathematics, and other fields, refer to App.~\ref{app:speech_instruction}.

\section{Related Work}
Speech contains rich non-semantic information compared to text~\citep{audiobench, bu2024roadmap, dynamic_superb}. For LLMs to achieve an accurate understanding of audio, they must have a comprehensive perception of speech rather than relying solely on text~\citep{ji2024wavchat, sdeval}. As a result, many researchers have studied how to build end-to-end speech LLMs~\citep{wavllm, SALMONN, qwen2audio, GAMA, fang2024llama, geng2025osum}. 

Some studies have found the \textit{less is more} phenomenon in LLMs with respect to data usage \citep{zhou2024lima,song-etal-2025-less}, meaning that efficient use of data can also achieve good performance. However, for speech LLMs, data efficiency has not been fully explored. Therefore, this work addresses this issue by focusing on the key challenge of speech-text alignment.

The acoustic features and text features differ significantly in both their representation space and length. To address this issue, \citet{qwen2audio, qwenaudio} employ convolution network to down-sample the speech, while others opt for solutions with more learnable parameters, such as Q-Former~\citep{SALMONN} and linear layers~\citep{wavllm}. Unlike previous work, the proposed Soundwave implements two adapters to address differences in representation and length, which also make training more efficient.

Speech LLMs are primarily designed for two capabilities: Speech and Sound. \citet{SALMONN,wavllm} combine Whisper with other feature extractors, such as BEATs~\citep{BEATs} and WavLM~\citep{WavLM}, to process sound features. 
\citet{qwen2audio} show that a fully fine-tuned encoder can also capture sound information. Our work demonstrates that a frozen encoder can efficiently process both types of features when provided with the proper data and training strategy.

\section{Conclusion}

Speech understanding is a core capability for multi-modal LLMs, yet current speech LLMs often rely on enormous amounts of training data, putting them out of reach for most academic researchers due to the high costs involved. To address this, we developed a more data-efficient solution: a three-stage training strategy paired with a model architecture that incorporates two adapters. This approach effectively tackles the mismatches in representation and length between speech and text. The trained Soundwave delivers top-tier performance on the AIR-Bench speech tasks,  while requiring significantly less training data.

\bibliography{custom} 
\bibliographystyle{nips}

\appendix

\section*{Limitations}
\raggedright

Our work still has some limitations, specifically in the following three aspects:

\begin{itemize}
    \item We have not verified the feasibility of our approach on larger models with more parameters.
    \item Due to time and manpower limitations, the amount of sound data we have labeled from the scene dataset is still relatively small. As a result, we are unable to conduct in-depth experiments to determine the optimal amount of sound data to include.
    \item Due to the lack of relevant data, our model does not perform well in music understanding tasks and has limited support for multiple languages.
 
\end{itemize}

Next, we will expand the parameter size of our model to verify the feasibility of our approach on larger models. We will also incorporate music understanding and multilingual data to enhance these capabilities. In addition, we will continue annotating the sound data to further validate the optimal data ratio. We also hope that other researchers in the community will conduct related studies.

\section*{Ethical Considerations}
\paragraph{Use of Artifacts} Ours study employs Whisper Large V3 as the audio encoder to extract and process speech input data and utilizes Llama-3.1-8B-Instruct as the foundation model for downstream tasks. In using these models, we adhere to academic standards and have cited their original papers and relevant documentation to ensure proper scholarly attribution. Additionally, Whisper is released under \href{https://opensource.org/license/mit}{MIT License}, while Llama-3.1-8B-Instruct is subject to \href{https://www.llama.com/llama3_1/license/}{Llama 3.1 Community License}. We have ensured that our application of the model does not violate any of the specified restrictions, thereby maintaining compliance with the license terms.

\paragraph{Data Collection} All the datasets used in our study are publicly released open-source datasets, and we strictly adhere to the corresponding open-source license agreements to ensure the legality and compliance of the data sources. In addition, the supplementary data annotation work we conducted did not involve any data privacy or sensitive information. Detailed procedures and workflow of the data annotation work can be found in Section \ref{sec:sound_process}. The content related to Statistics For Data can be found in Section \ref{app:data_description}.

\paragraph{Computational Experiment Design and Execution} In Section \ref{sec:Experiments Setting}, we detail the number of parameters of the base model used, the total computational budget, and the computing infrastructure employed. Tab.~\ref{tab:training_configuration} in Section \ref{training_setting} lists the hyperparameters used during training and other related configuration details. In Sections \ref{sec:Experiments Results} and \ref{Analysis}, we present the final training results and a comparative analysis of experiments.

\paragraph{Data Annotation and Ethical Compliance}In the Section \ref{sec:sound_process}, we provide detailed explanations of the manual data annotation work. Section \ref{sec:Data Splitting} details our data processing methods. We display the complete instruction text given to participants in Fig.~\ref{fig:annotation_page}, and explain our volunteer recruitment methods, salaries, and annotator characteristics in Section \ref{sec:Volunteer sources} and \ref{sec:Volunteer authorization}. Throughout the entire data processing procedure, no ethical risks to personal privacy or data security were posed, and therefore no ethics committee review was required. 

\paragraph{Use of AI Tools} In the course of this research project, AI tools were only utilized in specific aspects, such as assisting with coding and providing grammar checks and language refinement in the writing of the paper, to enhance efficiency and textual quality. Beyond these applications, the core research content, data processing, experimental design, analysis, and conclusions were all independently conducted by the research team, without any other form of artificial intelligence involvement, ensuring the rigor and originality of the study.

\section{Data Construction and Preparation}
\subsection{Sound Re-annotation}
\label{sec:sound_process}

Since a scene segment may contain multiple sounds, we allow users to select multiple sound labels. In the end, we labeled 7,863 audio files, totaling 7.30 hours, with 5,197 files being single-label, totaling 4.81 hours. Specific details can be found in the following sections. We will release all of these labeled data.

\subsubsection{Data Splitting}
\label{sec:Data Splitting}
We divide the original 10s data into combinations of 3s-3s-4s for annotation.

\subsubsection{Data Annotation}
\begin{figure}[th]
    \centering
    \includegraphics[width=0.6\linewidth]{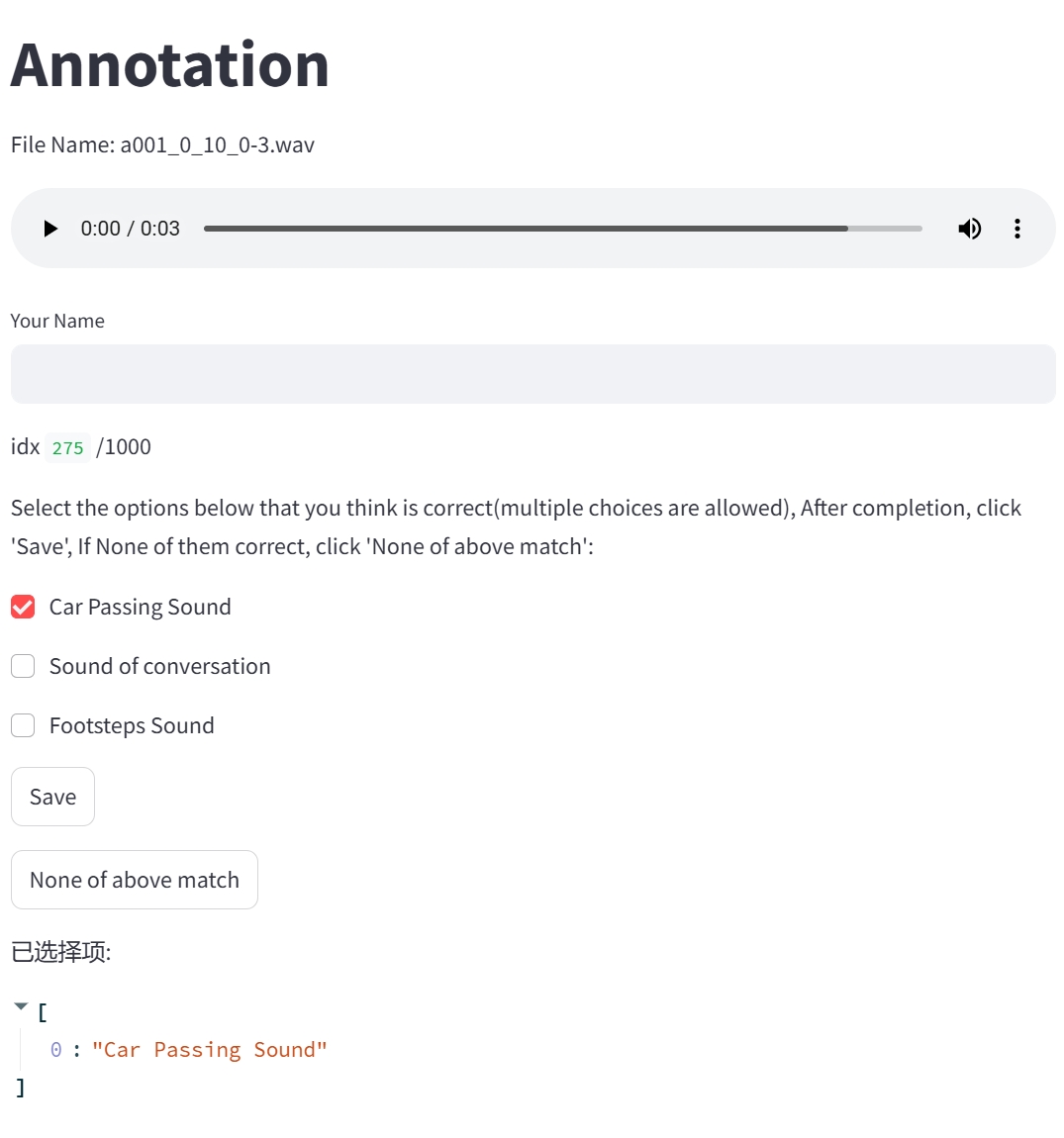}
    \caption{Page for sound annotation.}
    \label{fig:annotation_page}
\end{figure}

In accordance with the pre-existing scene labels, we established detailed sound annotations, such as waves and birdsong for a beach scenario. Subsequently, we recruited a number of volunteers to perform data annotation tasks. The interface utilized for this process is illustrated in Fig.~\ref{fig:annotation_page}.

\subsubsection{Volunteer Sources and Salaries}
\label{sec:Volunteer sources}
We extensively recruited volunteers for this project, comprising 70\% undergraduate students, 25\% graduate students, and 5\% individuals who have already graduated. Each participant was compensated with a one-time payment of 200 RMB, which aligns with the prevailing wage levels in mainland China.

\subsubsection{Volunteer Authorization}
\label{sec:Volunteer authorization}
All volunteers have agreed to the public release of their labeled data to promote academic research within the community.

\subsection{Data Process}
\label{app:data_process}
The process from raw data to final application in this paper includes two steps: Data Selection and Data Filtration.

\noindent\textbf{Data Selection}
 
We performed data selection on the following dataset:
\begin{itemize}
    \item \textbf{TED-LIUM}, we selected the speaker adaptation part as our dataset, as it is more balanced and representative in characteristics (number of speakers, gender, duration)~\citep{TED-LIUM}.
    \item \textbf{GiGaSpeech} It contains five data sizes: XL, L, M, S, and XS. We noticed that the WER limit for XL is relatively loose, so we did not choose this size of data. At the same time, due to the large number of data in L, it would cause GigaSpeech’s data to occupy too high a proportion, while the data in S and XS are too small. Therefore, we ultimately chose M.
    \item \textbf{Common Voice (En)}, 
    Common Voice contains multiple versions. Since we mainly use it to construct the data required for Covost2~\citep{wang2021covost}, we selected the matching version, which is Common Voice En 15.0. At the same time, we also use it to construct the data required for Age Prediction, which does not have specific version requirements. For convenience, we consistently use Common Voice En 15.0.
    \item \textbf{Common Voice (Ja)}, 
    Common Voice contains multiple versions. Since we are using this data to build a Language Identification task, and in order to balance with data from other languages, we need to select around 15 hours of data. Smaller versions do not provide enough data, and larger versions would result in unnecessary overhead, so we ultimately chose Common Voice ja 7.0.
\end{itemize}
\noindent\textbf{Data Filtration}
To ensure data quality, we performed filtering on the dataset:
\begin{itemize}
    \item \textbf{Duration}, it cannot exceed 30 seconds for the Whisper encoder, and it cannot be less than 3 seconds for more balanced training.
    \item \textbf{Lenth} 
    We excluded transcriptions that were longer than 200 to ensure the stability of the training process. We limit the frame length to no more than 100 times the text length, as exceeding this indicate the speech contains excessive noise.
    \item \textbf{WER} 
    To ensure stable training, we only retained the data with a WER of less than 10\% in the Whisper recognition results.
\end{itemize}

\noindent\textbf{Instruction Data Generation}
During the SFT data preparation phase, we used the GPT-4o-mini model to automatically generate question-answer pairs for all the training tasks. For example about the Fisher conversation ASR data, the process involved extracting dialogue segments from the ASR dataset and inputting them into the GPT-4o-mini model, which was instructed to generate contextually relevant questions framed from an objective third-person perspective. For each question, the model also generated corresponding answers directly derived from the conversation. These question-answer pairs were then incorporated into the training dataset for the chat task.
\subsection{Data Statistics}
\label{app:data_description}

We present the usage of audio data and the total amount for each dataset in Table~\ref{datasets}. It is important to note that in this statistical process, the same data is counted only once across different stages, only once across different tasks, and only once even if constructed using different methods within the same task. If you are interested in the specific data usage for each stage and task, please refer to the subsequent section.

\subsubsection{Alignment Stage Data}
In Tab.~\ref{tab:first_stage_datasets}, we present the datasets used during the alignment stage, along with their respective quantities and durations.

\begin{table}[h]
\centering
\caption{Summary of datasets, their size, and duration used in the alignment stage.}
\label{tab:first_stage_datasets}
\begin{tabular}{lrr}
\toprule
\textbf{Dataset}                           & \textbf{Size}       & \textbf{Duration (h)}  \\ \midrule
Multilingual LibriSpeech         & 984,559    & 4081.61       \\
GigaSpeech                        & 713,394    & 805.11        \\
TED-LIUM                          & 143,641    & 244.02        \\
TUT w./ libritts        & 4,698      & 18.48         \\
Vocalsound w./ libritts & 19,737     & 82.42         \\
Europral-ASR                         & 718,663    & 418.42        \\
LibriSpeech                       & 281,241    & 961.05        \\
TextrolSpeech                     & 170,502    & 246.85        \\
\textbf{SUM}                      & \textbf{3,036,435} & \textbf{6,857.97} \\ \bottomrule
\end{tabular}
\end{table}

As described in Section~\ref{sec:data_augmentation}, the sound and scene data used in this stage are both constructed with LibriTTS~\citep{lirbiTTS} inserted in between. "TUT w./ LibriTTS" refers to the dataset where TUT is sandwiched between LibriTTS, while "TUT w./ LibriTTS" refers to the dataset where VocalSound is sandwiched between LibriTTS.

\subsubsection{Shrinking Stage Data}
In Tab.~\ref{tab:summary_shrinking}, we present the datasets used during the shrinking stage, along with their respective quantities and durations.
\begin{table}[ht]
\centering
\caption{Summary of datasets, their size, and duration in the shrinking stage.}
\label{tab:summary_shrinking}
\begin{tabular}{lrr}
\toprule
\textbf{Dataset}                    & \textbf{Size}        & \textbf{Duration (h)}  \\ \midrule
ASR (First Stage)           & 3,012,000   & 6757.06       \\
Covost2 (en-de)             & 232,953     & 364.89        \\
MUST-C (en-de)              & 226,810     & 395.10        \\
Fisher                      & 153,102     & 1091.42       \\
TUT w./ LibriTTS            & 4,937       & 19.33         \\
VocalSound w./ LibriTTS     & 19,737      & 82.42         \\ 
\textbf{SUM}                & \textbf{3,649,539} & \textbf{8,710.23} \\ \bottomrule
\end{tabular}
\end{table}

Here, we have two points that need clarification: 1) Why is the ASR data reduced compared to the first stage? 2) Why has the "TUT w./ LibriTTS" data increased compared to the first stage?

Here, we will explain why the ASR data in this stage is reduced. In this stage, the LoRA parameters of the large model need to be trained. To ensure stable training, it is important to avoid overlap between tasks executed by different instructions. As shown in Figure 1, we have a dedicated task for Sound, which primarily involves predicting special Sound tokens. To prevent overlap, we removed the data containing Sound special tokens from the ASR dataset.

Here, we explain the reason for the increased amount of "TUT w./ LibriTTS" data in this stage. In the previous training phase, we removed data that was deemed to have a high level of noise during manual annotation in order to ensure the stability of the training process. However, in this phase, we included these data to enhance the model's robustness.

To further illustrate what our tasks are like, we have provided our text instructions and expected responses in Table~\ref{table:1st,2nd,example}. 
\begin{table}[ht]
\renewcommand{\arraystretch}{1.5} 
\centering
\small
\caption{Examples of input and output results from the first and second stages.} 
\begin{tabular}{c >{\raggedright\arraybackslash}m{4cm} >{\raggedright\arraybackslash}m{5.9cm}} 
\hline
 \textbf{Task} & \textbf{Instruction} & \textbf{Answer} \\ \hline
 \textbf{ASR} & \textit{I'd like to know what the person says in the recording.} & The recording indicates: please don't tease me so. \\ \hline
\textbf{English to German} & \textit{Please provide a professional translation of the attached English speech into German.} & The translated text is: Die Kreisstadt ist Phenix City. \\ \hline
\textbf{Chat} & \textit{What does the second person believe a friend would not ask you to do?} & A friend would not ask you to do something that might cause harm. \\ \hline
\textbf{Conversational ASR} & \textit{Transcribe the conversation between two persons.} & The first person says: Yeah, now that's something we have been toying with. We went to, uh, my son and I; he has a mowing business, he's sixteen. \textbackslash{}nThe second person says: Mhm. \textbackslash{}nThe first person says: \textless{}lipsmack\textgreater \ And, uh, I mean, it's a pretty good business now; we've got it up over fifty lawns. \\ \hline
\textbf{TUT w./ LibriTTS} & \textit{Please provide a complete transcription of this audio with the sound label.} & The complete transcription of this audio is: Matthew Cuthbert is surprised. \textless{}waves\textgreater \ In fact, he had looked at twenty very much as he looked at sixty, lacking a little of the grayness. \\ \hline
\end{tabular}

\label{table:1st,2nd,example} 
\end{table}

\subsubsection{SFT Data}

Our SFT data statistic details are shown in Table~\ref{tab:sft_data}.
To further clarify, we will specify in the following two paragraphs how we constructed the training data and tested these tasks.
\paragraph{Training Data}
Examples of the input and output for our SFT data can be found in Table \ref{tab:sft_data_example}. We employed two question-answer construction schemes: direct Q\&A and multiple-choice. However, some questions are not suitable for direct Q\&A. For instance, in tasks like Age Prediction, direct answers might confuse the model. Therefore, we only constructed multiple-choice format data for such tasks.

To enhance the instruction-following capability of our model, we do not rely on a single instruction to construct the training data. Instead, we use at least 50 different instructions for each task to build the dataset. Due to space limitations, we are unable to list all these instructions individually. We will subsequently release our complete training data.

\begin{table*}[ht]
\centering
\caption{Summary of datasets, their size, and duration
used in the SFT stage.}
\label{tab:sft_data}
\begin{tabular}{lccc}
\toprule
\textbf{Task} & \textbf{Dataset} & \textbf{Size} & \textbf{Duration (hours)} \\ 
\midrule
ASR& LibriSpeech & 281241 & 961.05 \\
Translation EN-DE &Covost2, MuST-C & 455648 & 753.18 \\
Speech Grounding &LibriSpeech & 23828 & 51.36 \\
Spoken Language Identification &Common Voice, Europal-ST & 269485 & 293.15 \\
Speaker Gender Recognition &TextrolSpeech  &319986 & 466.09 \\
Emotion Recognition & Four Datasets$^{*}$ &258541 & 347.18 \\
Speaker Age Prediction & Common Voice &77239 & 120.04 \\
Speech Entity Recognition & SLURP& 114792 & 91.62 \\
Intent Classification&  SLURP & 212473 &151.52 \\
Speaker Num. Verification &VoxCeleb & 156033 & 435.17 \\
Synthesized Voice Detection &FoR& 53696 & 47.55 \\
Vocal Sound Classification & VocalSound &29962 & 34.77 \\
Acoustic Scene Classification & TUT2017, CochlScene& 159590 & 443.31 \\
Chat & Fisher &132209 & 956.25 \\ 
Speech Instruction& AnyInstruct & 106770&206.30  \\ 
\textbf{SUM}&\textbf{18}&\textbf{2651493}&\textbf{5358.54}\\

\bottomrule
\end{tabular}
* These four datasets include: TextrolSpeech, CREMA-D, IEMOCAP, MELD.
\end{table*}

\begin{table}[]
\scriptsize
\caption{SFT Data input-output pair examples.}
\label{tab:sft_data_example}
\begin{tabular}{@{}l p{8.8cm} p{3.5cm}@{}}
\toprule
\textbf{Task}                                                   & \textbf{Instruction example}                                                                                                                                                                                                                                                                                                                                                                                                          & \textbf{Answer}                                                                                                                                                        \\ \midrule
SG              & \textit{Choose the most suitable answer from options A, B, C, and D to respond the question in next line, Please think step by step and you may only choose A or B or C or D.\textbackslash{}nIdentify the time period in which 'the' is prominently featured in the audio.\textbackslash{}nA. {[}4.60, 5.09{]}\textbackslash{}nB. {[}3.72, 3.86{]}\textbackslash{}nC. {[}5.21, 5.29{]}\textbackslash{}nD. {[}1.21, 1.34{]}} & Based on the position of the word 'the' in the sentence, the most likely choice is C. {[}5.21, 5.29{]}.                                                       \\ \hline
SLI & \textit{Which language is used by the speaker in this recording?}                                                                                                                                                                                                                                                                                                                                                            & The language used by the speaker in this recording is de.                                                                                                     \\  \cline{2-3}
                                & \textit{Choose the most suitable answer from options A, B, C, and D to respond the question in next line, you may only choose A or B or C or D.\textbackslash{}nWhat language is being used for the speaker in this speech?\textbackslash{}nA. it\textbackslash{}nB. zh-CN\textbackslash{}nC.en\textbackslash{}nD. ja}                                                                                                       & The most suitable answer from options is B. zh-CN.                                                                                                            \\
\hline
SGR     & \textit{Can you identify whether the speaker in this audio is male or female?}                                                                                                                                                                                                                                                                                                                                               & The speaker in the audio is identified as female.                                                                                                             \\ \cline{2-3}
                                & \textit{Choose the most suitable answer from options A, B, C, and D to respond the question in next line, you may only choose A or B or C or D.\textbackslash{}nIs it possible to determine the gender of the speaker by analyzing the voice traits in this audio?\textbackslash{}nA. male\textbackslash{}nB. female\textbackslash{}nC.None\textbackslash{}nD. None}                                                         & The most suitable answer from options is A. male.                                                                                                             \\ \hline
ER            & \textit{What emotion is the speaker's tone permeated with?}                                                                                                                                                                                                                                                                                                                                                                  & The speaker's tone is permeated with sadness.                                                                                                                 \\ \cline{2-3}
                                & \textit{What emotion can be sensed from the speaker's message?\textbackslash{}nChoose the option that best fits from A, B, C, and D.\textbackslash{}nA: joy\textbackslash{}nB: surprise\textbackslash{}nC: neutral\textbackslash{}nD: disgust}                                                                                                                                                                               & The option that best fits is C: neutral.                                                                                                                      \\ \hline
AP         & \textit{Choose the most suitable answer from options A, B, C, and D to respond the question in next line, you may only choose A or B or C or D.\textbackslash{}nWhich age group's characteristics are evident in the speaker's voice?\textbackslash{}nA. thirties to fourties\textbackslash{}nB. seventies to eighties\textbackslash{}nC.fifties to sixties\textbackslash{}nD. teens to twenties}                            & The most suitable answer from options is A. thirties to fourties.                                                                                             \\  \hline
SER      & \textit{What is the earliest audible reference to 'food\_type' in this recording?}                                                                                                                                                                                                                                                                                                                                           & The earliest audible reference to 'food\_type' in this recording is 'chinese'.                                                                                \\  \cline{2-3}
                                & \textit{Choose the most suitable answer from options A, B, C, and D to respond the question in next line, you may only choose A or B or C or D.\textbackslash{}nIdentify the initial 'person' term mentioned in this recording.\textbackslash{}nA. send\textbackslash{}nB. about\textbackslash{}nC.email\textbackslash{}nD. nancy}                                                                                           & The most suitable answer from options is D. nancy.                                                                                                            \\  \hline
IC          & \textit{What do you think the speaker's message is intended to be in this audio?}                                                                                                                                                                                                                                                                                                                                            & Based on what the speaker mentions: 'siri what is one american dollar in japanese yen', the speaker's message in this audio is intended to be 'qa\_currency'. \\  \cline{2-3}
                                & \textit{Choose the most suitable answer from options A, B, C, and D to respond the question in next line, you may only choose A or B or C or D.\textbackslash{}nWhat do you think the speaker aims to achieve with their message?\textbackslash{}nA. general\_praise\textbackslash{}nB. transport\_ticket\textbackslash{}nC.locations\textbackslash{}nD. general\_quirky}                                                    & The most suitable answer from options is A. general\_praise.                                                                                                  \\  \hline

SNV    & \textit{Choose the most suitable answer from options A, B, C, and D to respond the question in next line, you may only choose A or B or C or D.\textbackslash{}nWhat is the total count of speakers involved in this speech?\textbackslash{}nA. 2\textbackslash{}nB. 4\textbackslash{}nC.3\textbackslash{}nD. 1}                                                                                                             & The most suitable answer from options is A. 2.                                                                                                                \\  \hline
SVD    & \textit{Based on your assessment, is this speech Real or Fake?}                                                                                                                                                                                                                                                                                                                                                              & The speech is Fake.                                                                                                                                           \\ \bottomrule
\end{tabular}
SG is for Speech Grounding; SLI is for Spoken Language Identification; SGR is for Speaker Gender Recognition; ER is for Emotion Recognition; AP is for Age Prediction; SER is for Speech Entity Recognition; IC is for Intent Classification; SNV is for Speaker Number Verification; SVD is for Synthesized Voice Detection.
\end{table}

\subsection{Mitigating Data Leakage Risks}
\label{app:data_leak}
In this section, we will discuss the risks of data leakage in several parts. For one set of tasks, we used non-homogeneous training data, while for another set, although we employed homogeneous data, we rigorously considered the issue of data leakage.
\subsubsection{Non-homogeneous Training Data}
\textbf{Speech Gender Recognition} The task involved in the test set is AIR-Bench\citep{airbench}, which uses Common Voice\citep{ardila2019common_voice} and MELD\citep{MELD} to construct the data. We use TextrolSpeech\citep{TextrolSpeech} to construct the data, which is considered non-homogeneous data in comparison.

\noindent\textbf{Spoken Language Identification} 
This task involves a total of 7 languages: Chinese, English, Italian, German, French, Spanish, and Japanese. AIR-Bench~\citep{airbench} uses Covost2~\citep{wang2021covost} in its construction, which is sourced from Common Voice~\citep{ardila2019common_voice}. The construction of data in English, Italian, German, French, and Spanish, we used Europarl-ASR~\citep{europarlASR}, while for Chinese data, we used AISHELL3~\citep{aishell3}. These sources are different from Common Voice, so there is no data leakage. For Japanese data construction, we used Common Voice, which is the same source as AIR-Bench, so we paid special attention to potential leakage issues. We noticed that there were only two Japanese samples in AIR-Bench, so we manually removed these two entries.

\subsubsection{Homogeneous Training Data}
\textbf{Speech Grounding}
Since our training set is constructed using the same dataset as in the AIR-Bench test, we made sure that the test set was not included in the training. We removed data where the same word in the same position was queried in the audio. Specifically, due to the difficulty of ensuring that randomly selected data doesn't overlap during the selection process, we adopted a post-processing approach where we deleted training data with the same filename and identical queries.

\noindent\textbf{Emotion Recognition}
During the construction of our training dataset, we utilized the TextrolSpeech~\citep{TextrolSpeech}, RAVDESS~\citep{RAVDESS}, CREMA-D\citep{Crema-d}, IEMOCAP~\citep{IEMOCAP}, and MELD~\citep{MELD} datasets. Notably, TextrolSpeech is composed of multiple datasets, including ESD, MEAD, MESS, SAVEE, and TESS. Given that AIR-Bench incorporates data from IEMOCAP and MELD, we have entirely excluded these datasets from our training data. Additionally, the test set of MELD has also been removed to ensure data integrity and prevent potential data leakage.

\noindent\textbf{Speech Entity Recognition and Intent Classification}
We used the same source data as AIR-Bench for construction, both utilizing SLURP~\citep{SLURP}, so we paid special attention to data leakage issues. Since AIR-Bench retained the original file names for all its files, we directly removed this portion of the data from our dataset.

\noindent\textbf{Speaker Number Verification} 
We used the same source data as AIR-Bench for construction, both utilizing VoxCeleb~\citep{VoxCeleb}, 
We used a fully random selection method, choosing a series of speech pairs to form our training set. Since the random selection process is hard to control, we removed any data that had already appeared in AIR-Bench after the selection.

\noindent\textbf{Synthesized Voice Detection} 
We used FoR~\citep{FoR} to construct our training set. We noticed that FoR overlaps with AIR-Bench. However, AIR-Bench does not provide detailed records of the specific sources of these data, so we removed the repeated test and development sets from FoR.

\subsection{Dataset Lisence}

The paper and license for the dataset we used are listed in Table~\ref{tab:license_and_cite}. There are several points regarding the usage of data that need to be clarified.

\section{Training Configurations}
\label{training_setting}
The training settings of different stages are shown in Tab.~\ref{tab:training_configuration}. For all training and decoding processes, we set \textit{‘You are a helpful language and speech assistant. You are able to understand the speech content that the user provides and assist the user with a variety of tasks using natural language.’} as the system prompt.
\begin{table}[ht]
\centering
\caption{Overview of training parameters at different stages.}
\label{tab:training_configuration}
\begin{tabular}{lccc}
\toprule
\textbf{Settings} & \textbf{Stage 1} & \textbf{Stage 2} & \textbf{Stage 3} \\ 
\midrule
Batch &32&16&8\\
Learning rate &1e-4&3e-5&3e-5\\
Accumulation steps &8&8&4\\
Training param. &144M&266M&122M\\
\bottomrule
\end{tabular}
\end{table}

\begin{figure}[th]
    \centering
    \includegraphics[width=0.48\textwidth]{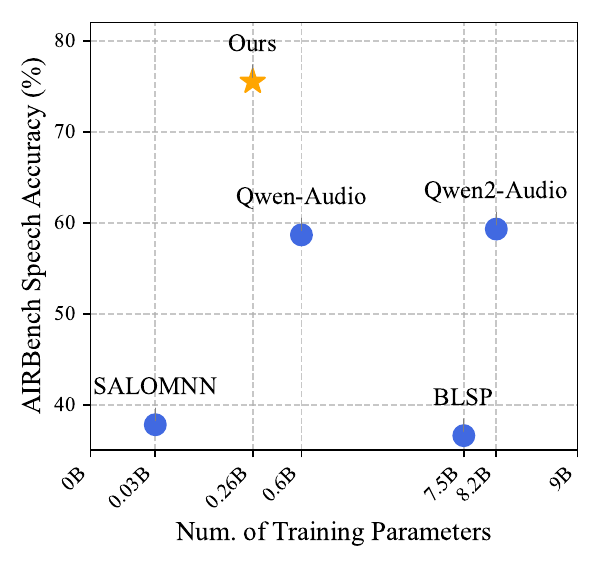} 
    \caption{AIR-Bench speech accuracy with number of training parameters.} 
    \label{fig:para_vs_acc} 
\end{figure}

All in all, we trained a total of 266M parameters in our three-stage process. To better highlight the advantages of our model's parameters compared to others, we plotted the relationship between AIR-Bench performance and training parameters, which is shown in Fig.~\ref{fig:para_vs_acc}. 

\section{Instructions used in the experiment}

Table~\ref{tab:test_instructions} presents the instructions we used in the experiment on "Performance on Foundation Tasks". In order to align with the previous test results, we have fully adopted the instructions from its test data for AIRBench.

\begin{table*}[ht]
\centering
\caption{Instructions used in evaluation experiments}
\begin{tabular}{lp{10cm}}
\toprule
\textbf{Task} & \textbf{Instruction} \\
\midrule
\textbf{ASR} & What does the person say? \\
\textbf{SER} & What emotion does the speaker express in their tone? \\
          & Choose just one answer from A, B, C, or D. \\
          & A: neutral \textcolor{gray}{\textbackslash n}B: joy \textcolor{gray}{\textbackslash n}C: sadness \textcolor{gray}{\textbackslash n}D: disgust \\
\textbf{VSC} & Where's the sound in this clip originating from? \\
\textbf{ST} & What is the target language$^{*}$ translation of this English speech? \\
\bottomrule
\end{tabular}
\label{tab:test_instructions}
\\ 
* The target language includes: German, Dutch, Italian, Romanian, Spanish.
\end{table*}

\begin{table*}[ht]
\centering
\caption{The summary  for the dataset}
\label{tab:license_and_cite}
\begin{tabular}{ccc}
\toprule
\textbf{DataSet Name} & \textbf{Citation} & \textbf{License} \\
\midrule
Librispeech w./ timestamp& \citet{librispeech_asr_test_clean_word_timestamp}&\href{https://choosealicense.com/licenses/apache-2.0/}{Apache 2.0}\\
Librispeech &\citet{LibriSpeech} & \href{http://creativecommons.org/licenses/by/4.0/}{CC BY 4.0} \\
CREMA-D &\citet{Crema-d} & \href{http://opendatacommons.org/licenses/dbcl/1.0/}{DbCL-1.0} \\
TED-LIUM &\citet{TED-LIUM} & \href{https://creativecommons.org/licenses/by-nc-nd/3.0/}{CC BY-NC-ND 3.0} \\
MLS &\citet{MLS} & \href{https://creativecommons.org/licenses/by/4.0/}{CC BY 4.0} \\
Europarl-ASR & \citet{europarlASR}& \href{http://creativecommons.org/licenses/by/4.0/}{CC BY 4.0} \\
TextrolSpeech &\citet{TextrolSpeech} & \href{https://opensource.org/license/mit}{MIT License} \\
LibriTTS &\citet{lirbiTTS}&\href{http://creativecommons.org/licenses/by/4.0/}{CC BY 4.0}\\
VCTK &\citet{VCTK}&\href{https://opendatacommons.org/licenses/by/1.0/}{ODC-By 1.0}\\ 
TUT2017 &\citet{TUTscene} & \href{https://zenodo.org/records/400515}{Non-Commercial} \\
VocalSound &\citet{VocalSound} & \href{https://creativecommons.org/licenses/by-sa/4.0}{CC BY-SA 4.0} \\
MUST-C &\citet{MuST-C} & \href{https://creativecommons.org/licenses/by-nc-nd/4.0/}{CC BY-NC-ND 4.0} \\
Europarl-ST &\citet{Europarl-ST} & \href{https://creativecommons.org/licenses/by-nc/4.0/}{CC BY-NC 4.0} \\
Common Voice & \citet{ardila2019common_voice}& \href{https://creativecommons.org/publicdomain/zero/1.0/}{CC0 1.0 Universal} \\
CochlScene & \citet{CochlScene}& \href{https://creativecommons.org/licenses/by-sa/3.0/deed.en}{CC BY-SA} \\
SLURP &\citet{SLURP} & \href{https://creativecommons.org/licenses/by-nc/4.0/}{CC BY-NC 4.0} \\
RAVDESS & \citet{RAVDESS}& \href{https://creativecommons.org/licenses/by-nc-sa/4.0/}{CC BY-NC-SA 4.0} \\
IEMOCAP &\citet{IEMOCAP} & \href{https://sail.usc.edu/iemocap/Data_Release_Form_IEMOCAP.pdf}{IEMOCAP License} \\
MELD & \citet{MELD}& \href{https://www.gnu.org/licenses/gpl-3.0.html}{GPL-3.0} \\
Gigaspeech & \citet{gigaSpeech}& \href{https://choosealicense.com/licenses/apache-2.0/}{Apache 2.0} \\
Covost2 & \citet{wang2021covost}& \href{https://creativecommons.org/publicdomain/zero/1.0/}{CC0 1.0 Universal} \\
VoxCeleb & \citet{VoxCeleb}& \href{http://creativecommons.org/licenses/by/4.0/}{CC BY 4.0} \\
WaveFake & \citet{WaveFake}& \href{https://opensource.org/license/mit}{MIT License} \\
FoR & \citet{FoR}& \href{https://www.gnu.org/licenses/lgpl-3.0.html}{LGPL-3.0} \\
Fisher & \citet{cieri2004fisher}& 
\href{https://catalog.ldc.upenn.edu/license/ldc-non-members-agreement.pdf}{LDC License} \\
\bottomrule
\end{tabular}
\end{table*}

\begin{table*}[t]
    \centering
    \scriptsize
    \caption{Performance on the AIR-Bench sound and music foundation tasks.}
    \label{Airbench_sound_music}
    \setlength{\tabcolsep}{1.0mm}{
    \begin{tabular}{lcccccccccc}
        \toprule
        \multirow{2}{*}{\textbf{Task}} & \multirow{2}{*}{\textbf{Soundwave}} & \multirow{2}{*}{ \textbf{Qwen2-Audio}} & {\textbf{Qwen-Audio}} & \multirow{2}{*}{\textbf{SALMONN}} & \multirow{2}{*}{\textbf{BLSP}} & \multirow{2}{*}{\textbf{NExT-GPT}} & \multirow{2}{*}{\textbf{PandaGPT}} &  \\
        & & & \textbf{Turbo} & & & & &  \\
        \midrule 
        Audio Grounding & 23.1 & 34.9 & 41.6 & 24.0 & 34.6 & \textbf{62.2} & 38.3 &   \\ 
        Vocal Sound Classification & \textbf{91.7} & 89.3 & 78.1 & 45.3 & 29.8 & 23.5 & 31.6 &  \\
        Acoustic Scene Classification & \textbf{83.8} & 67.4 & 61.3 & 34.1 & 25.2 & 24.1 & 55.7 &   \\
        Sound Question Answering & 49.7 & \textbf{68.8} & 62.8 & 28.4 & 36.1 & 18.8 & 48.7 &  \\ 
        Sound avg. & 62.1 & \textbf{65.1} & 61.0 & 33.0 & 31.4 & 32.2 & 43.6 &  \\
        \midrule 
        Music Instruments Classification & 37.1 & \textbf{65.8} & 59.6 & 41.3 & 22.8 & 24.3 & 47.7 &  \\
        Music Genre Classification & 49.5 & \textbf{78.8} & 77.1 & 45.3 & 26.1 & 28.1 & 39.8 &  \\ 
        Music Note Analysis-Pitch & 27.7 & 28.7 & \textbf{30.1} & 26.4 & 23.5 & 25.1 & 26.4 &  \\
        Music Note Analysis-Velocity & 23.2 & 26.2 & 25.1 & 22.8 & 24.9 & 23.1 & \textbf{27.2} &  \\
        Music Question Answering & 65.0 & \textbf{65.7} & 62.5 & 54.6 & 31.0 & 47.1 & 50.7 &  \\
        Music Emotion Detection & 38.3 & \textbf{46.9} & 39.0 & 32.2 & 28.3 & 25.4 & 36.7 &  \\ 
        Music avg. & 40.1 & \textbf{52.0} & 48.9 & 37.1 & 26.1 & 28.9 & 38.1 &  \\
        \bottomrule
    \end{tabular}
    }
\end{table*}

\begin{table*}[t]
    \centering
    \scriptsize
     \caption{Performance on the AIR-Bench chat tasks.}
    \label{Airbench_chat}
    \setlength{\tabcolsep}{1.0mm}{
    \begin{tabular}{lcccccccc}
        \toprule
        \multirow{2}{*}{\textbf{Task}} & \multirow{2}{*}{\textbf{Soundwave}} & \multirow{2}{*}{ \textbf{Qwen2-Audio}} & { \textbf{Qwen-Audio}} & \multirow{2}{*}{\textbf{SALMONN}} & \multirow{2}{*}{\textbf{BLSP}} & \multirow{2}{*}{\textbf{NExT-GPT}} & \multirow{2}{*}{\textbf{PandaGPT}} &\textbf{Whisper}  \\
        & & & \textbf{Turbo} & & & & & \textbf{+GPT-4} \\
        \midrule 
        \textbf{Speech} & 6.41 & 7.18 & 7.04 & 6.16 & 6.17 & 3.86 & 3.58 & 7.54 \\
        \textbf{Sound} & 5.33 & 6.99 & 6.59 & 6.28 & 5.55 & 4.76 & 5.46 & / \\
        \textbf{Music} & 5.10 & 6.79 & 5.98 & 5.95 & 5.08 & 4.18 & 5.06 & / \\
        \textbf{Mixed Audio} & 4.98 & 6.77 & 5.77 & 6.08 & 4.52 & 2.92 & 2.93 & / \\
        \textbf{Average} & 5.46 & 6.93 & 6.34 & 6.11 & 5.33 & 4.13 & 4.25 & / \\

        \bottomrule
    \end{tabular}
    }

\end{table*}

\section{Other AIR-Bench Evaluation}

\subsection{AIR-Bench Sound and Music Foundation Tasks}
The AIR-Bench sound and music tasks evaluate models on various auditory capabilities. Sound tasks focus on identifying, classifying, and reasoning with environmental sounds, while music tasks involve classifying musical elements, analyzing pitch and velocity, and understanding emotional content. As Table~\ref{Airbench_sound_music} shown, Soundwave demonstrates exceptional performance in vocal sound and acoustic scene classification, achieving impressive accuracy. Though few sound data is used, sound average score of Soundwave still ranks second. This highlights its strong performance across sound-related tasks, even with limited data.
\subsection{AIR-Bench Chat Tasks}
The AIR-Bench chat tasks evaluate language models' ability to generate conversational responses based on speech, music, environmental sounds, and mixed audio. Evaluation is conducted using GPT-4-0125-preview, which rates the model's responses on accuracy, relevance, usefulness, and comprehensiveness, on a scale from 1 to 10. The evaluation results are shown in Table~\ref{Airbench_chat}.

\section{Speech Instruction}
\label{app:speech_instruction}
We demonstrate our model's ability to follow voice commands from two aspects: generation tasks and knowledge question-answering tasks.
\subsection{Generation Tasks}
\begin{figure*}[th]
    \centering
    \includegraphics[width=0.7\linewidth]
    {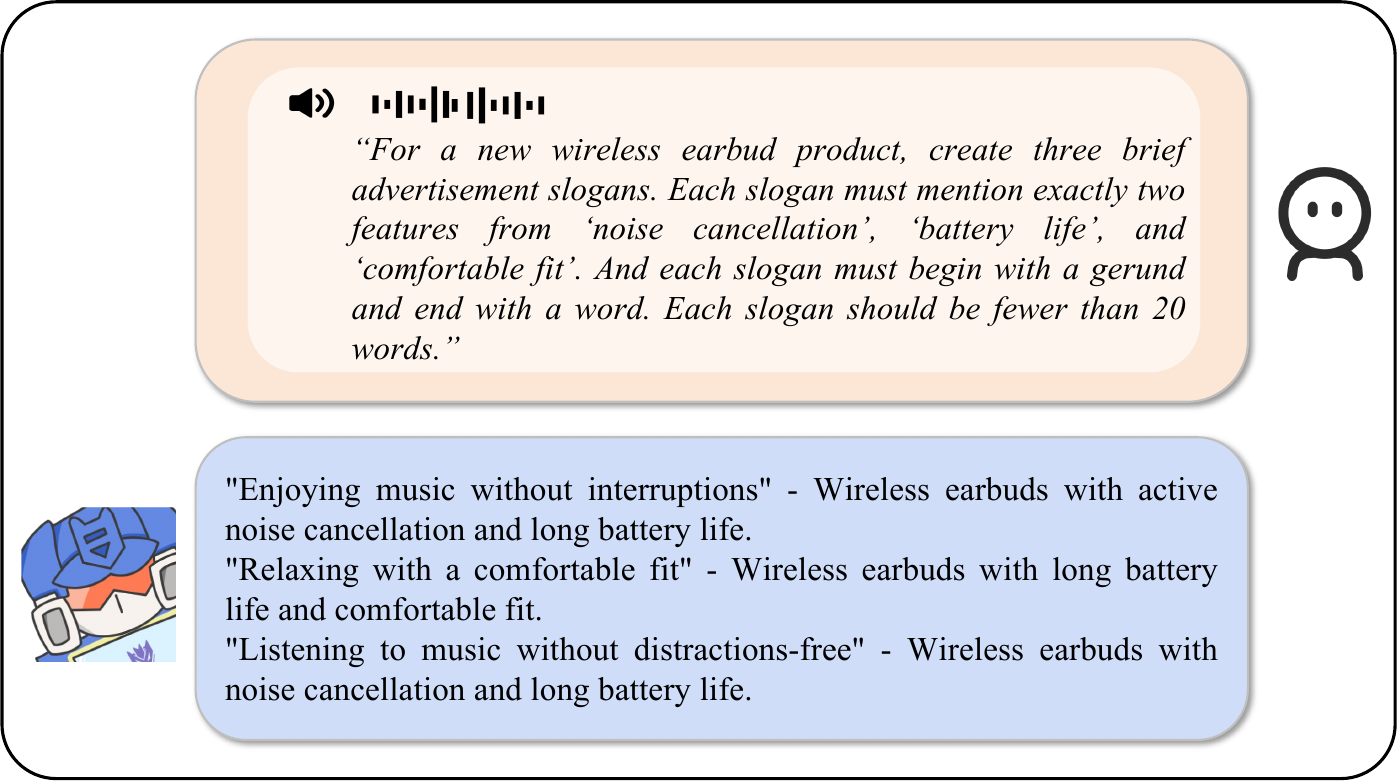}
    \caption{Generate task show case 1.}
    \label{fig: generate_case1}
\end{figure*}

\begin{figure*}[h]
    \centering
    \includegraphics[width=0.7\linewidth]
    {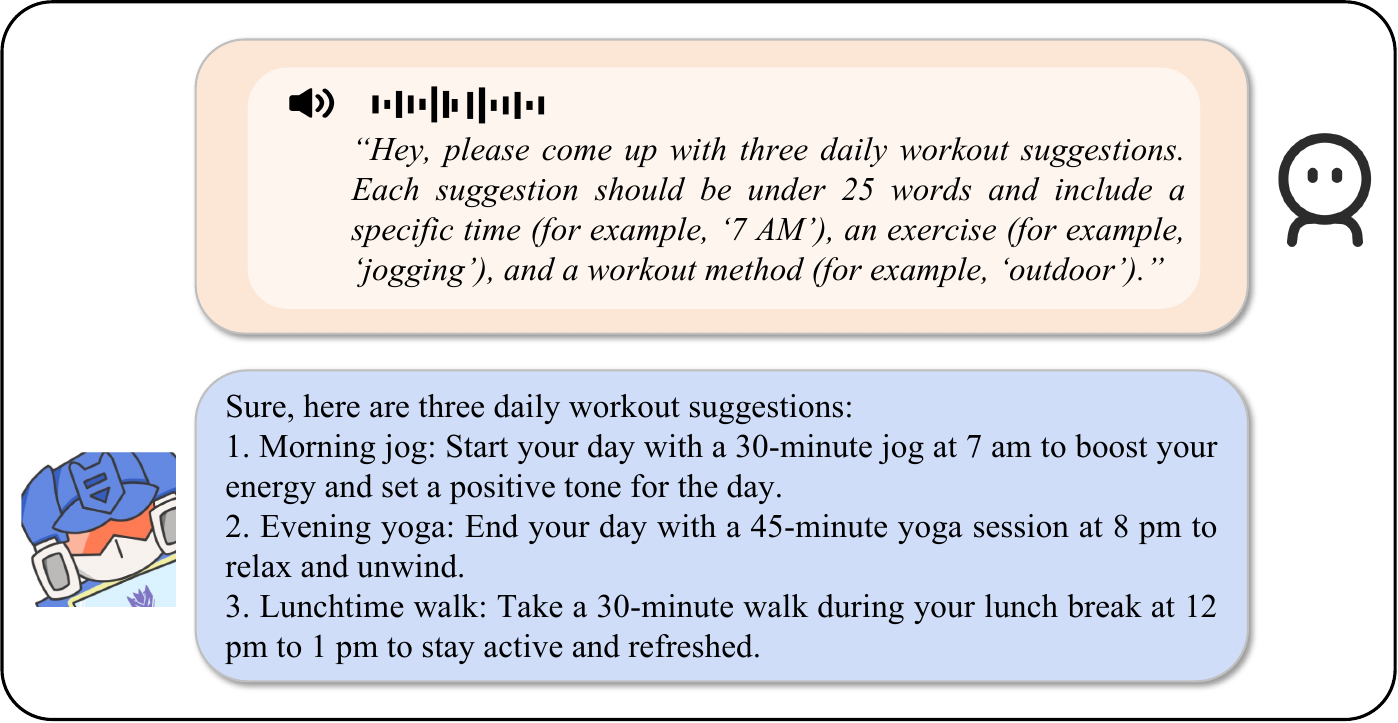}
    \caption{Generate task show case 2.}
    \label{fig: generate_case2}
\end{figure*}

\begin{figure*}[t]
    \centering
    \includegraphics[width=0.7\linewidth]
    {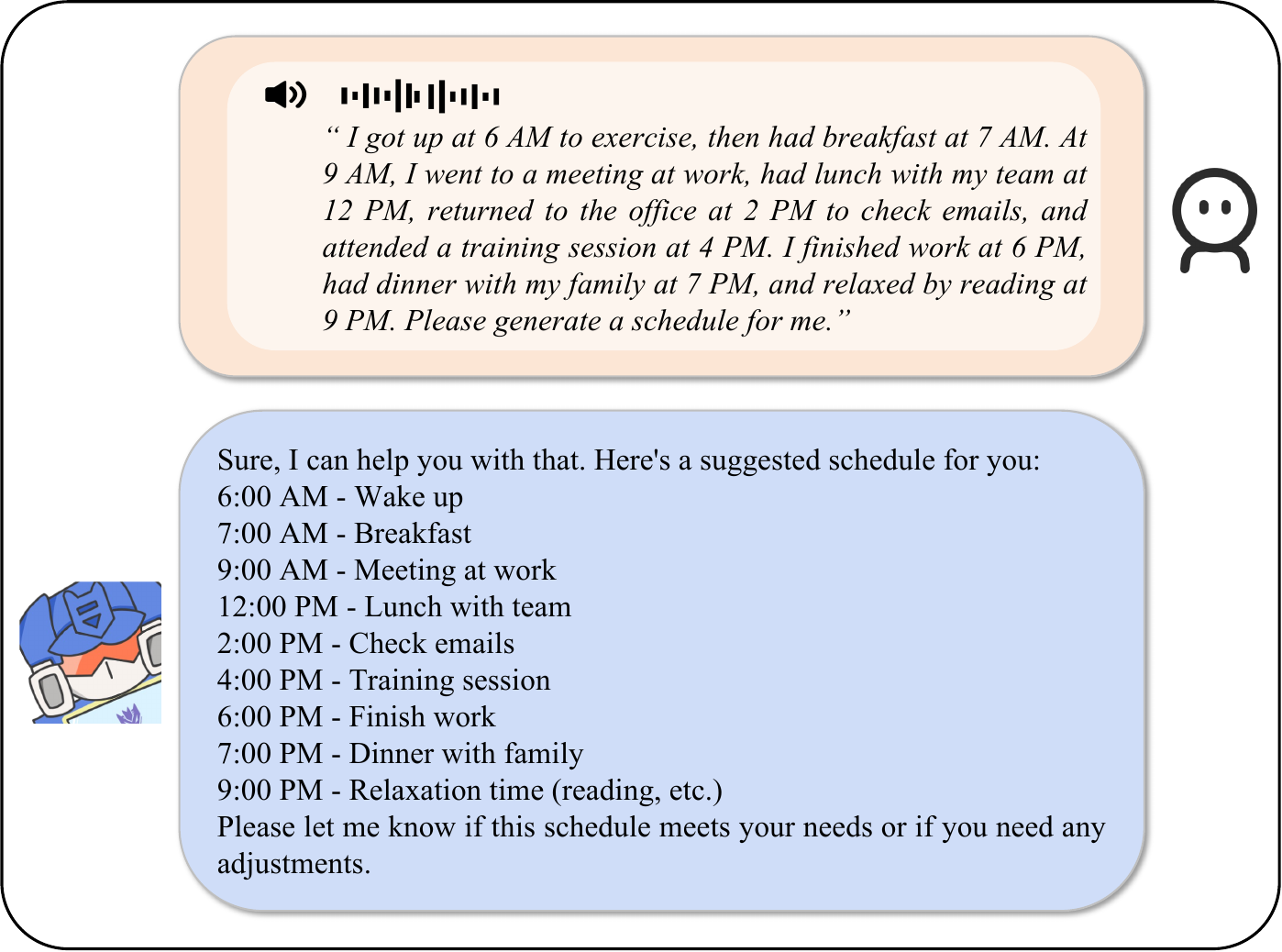}
    \caption{Generate task show case 3.}
    \label{fig: generate_case3}
\end{figure*}

We selected some commonly used generation tasks in daily life, which demonstrate our model's ability to assist in handling everyday affairs. Our presentation results are shown in Figures \ref{fig: generate_case1}, \ref{fig: generate_case2}, and \ref{fig: generate_case3}.

\subsection{Knowledge-based QA Tasks}

\begin{figure*}[t]
    \centering
    \includegraphics[width=0.7\linewidth]
    {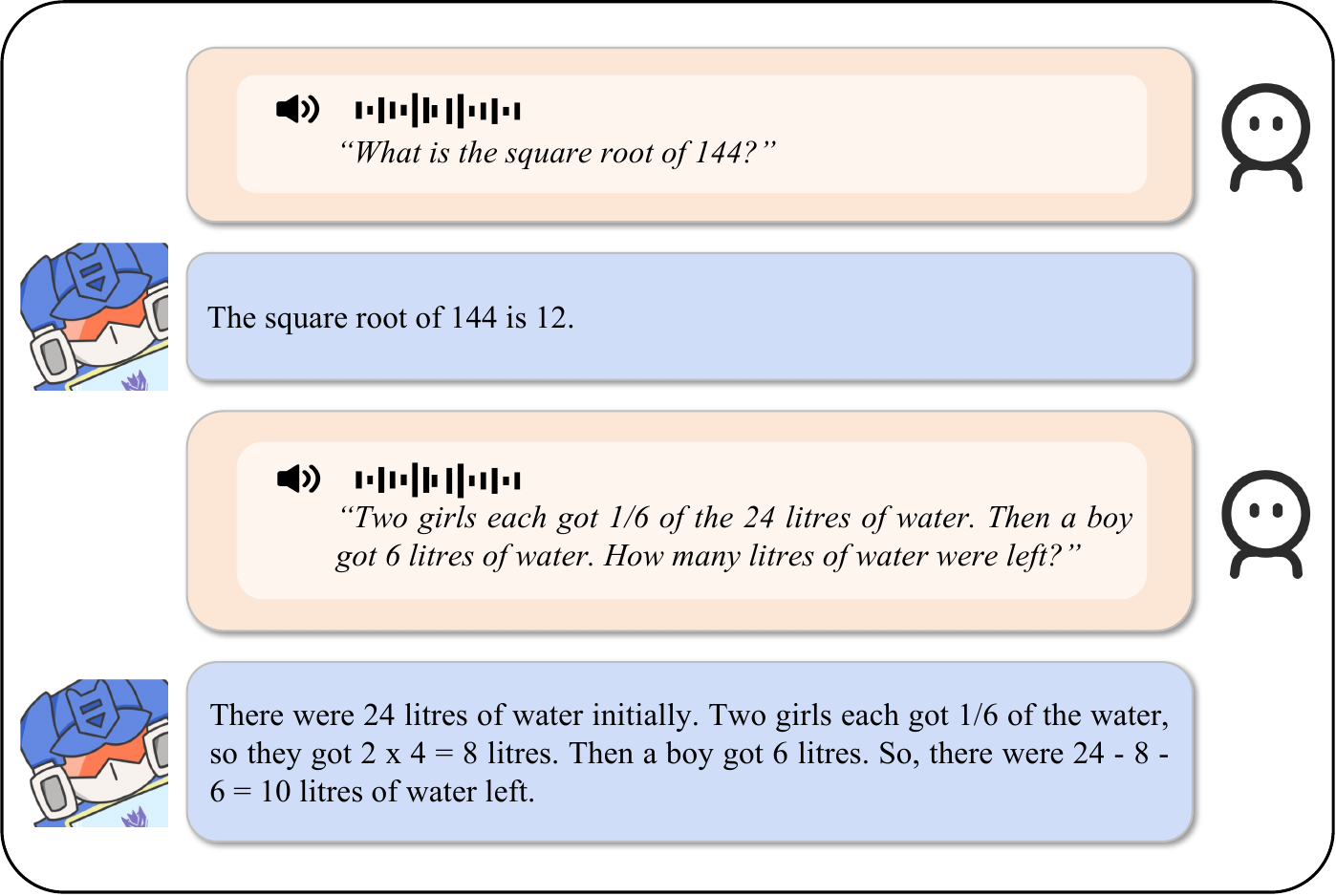}
    \caption{Knowledge-based QA about math.}
    \label{fig: math_case}
\end{figure*}

\begin{figure*}[t]
    \centering
    \includegraphics[width=0.7\linewidth]
    {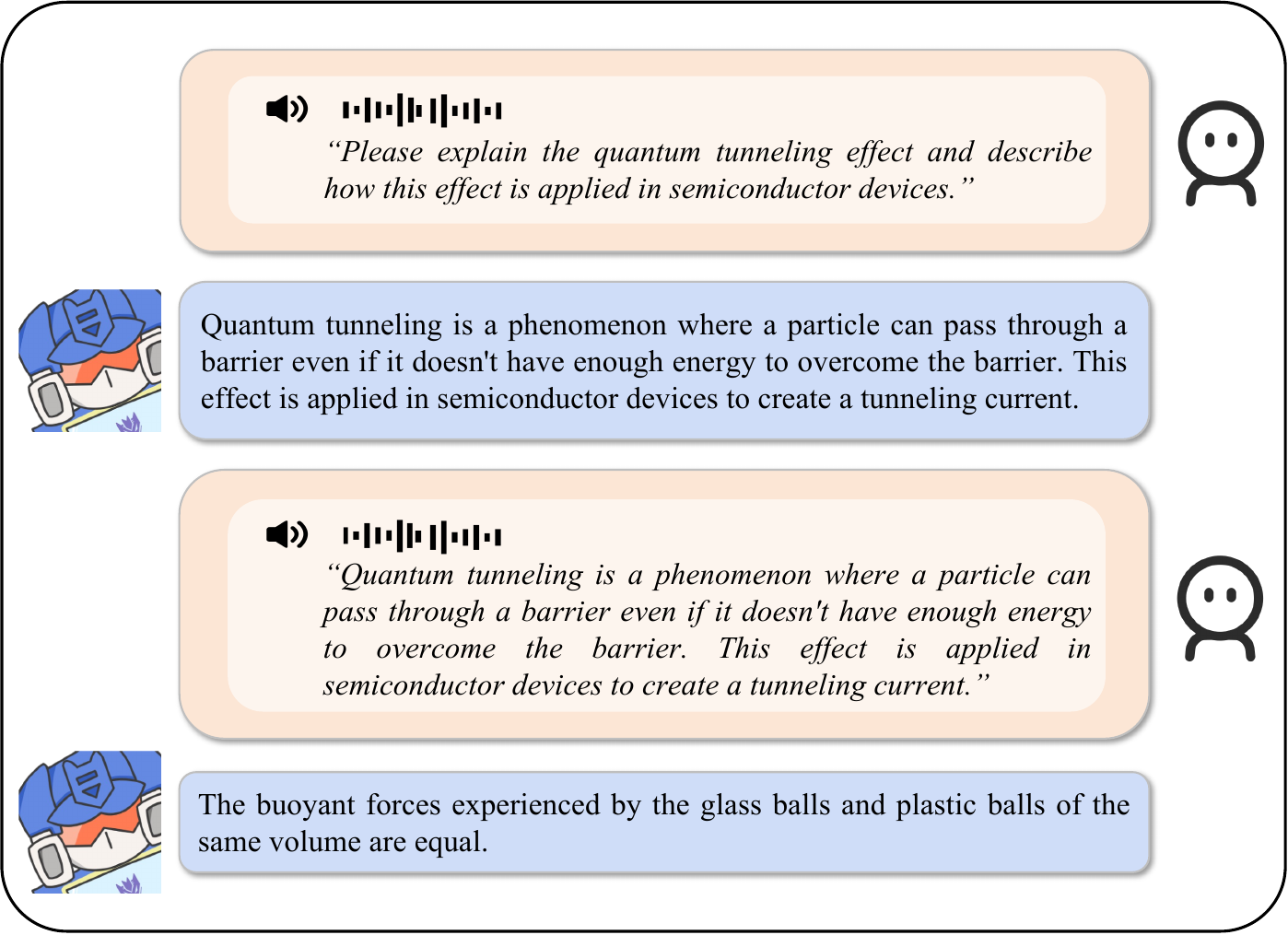}
    \caption{Knowledge-based QA about physics.}
    \label{fig: physics_case}
\end{figure*}

\begin{figure*}[t]
    \centering
    \includegraphics[width=0.7\linewidth]
    {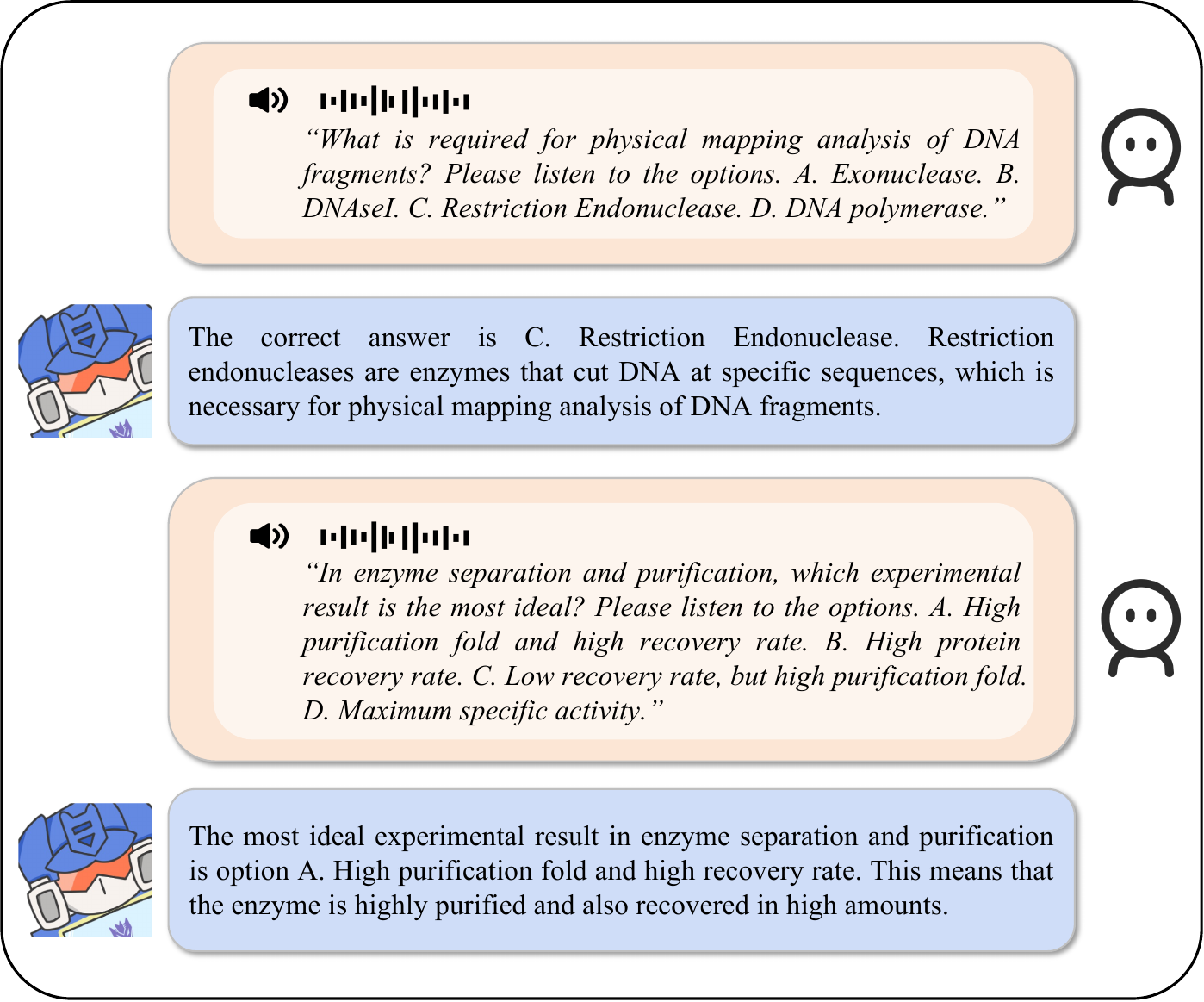}
    \caption{Knowledge-based QA about biology.}
    \label{fig: biology_case}
\end{figure*}

\begin{figure*}[t]
    \centering
    \includegraphics[width=0.7\linewidth]
    {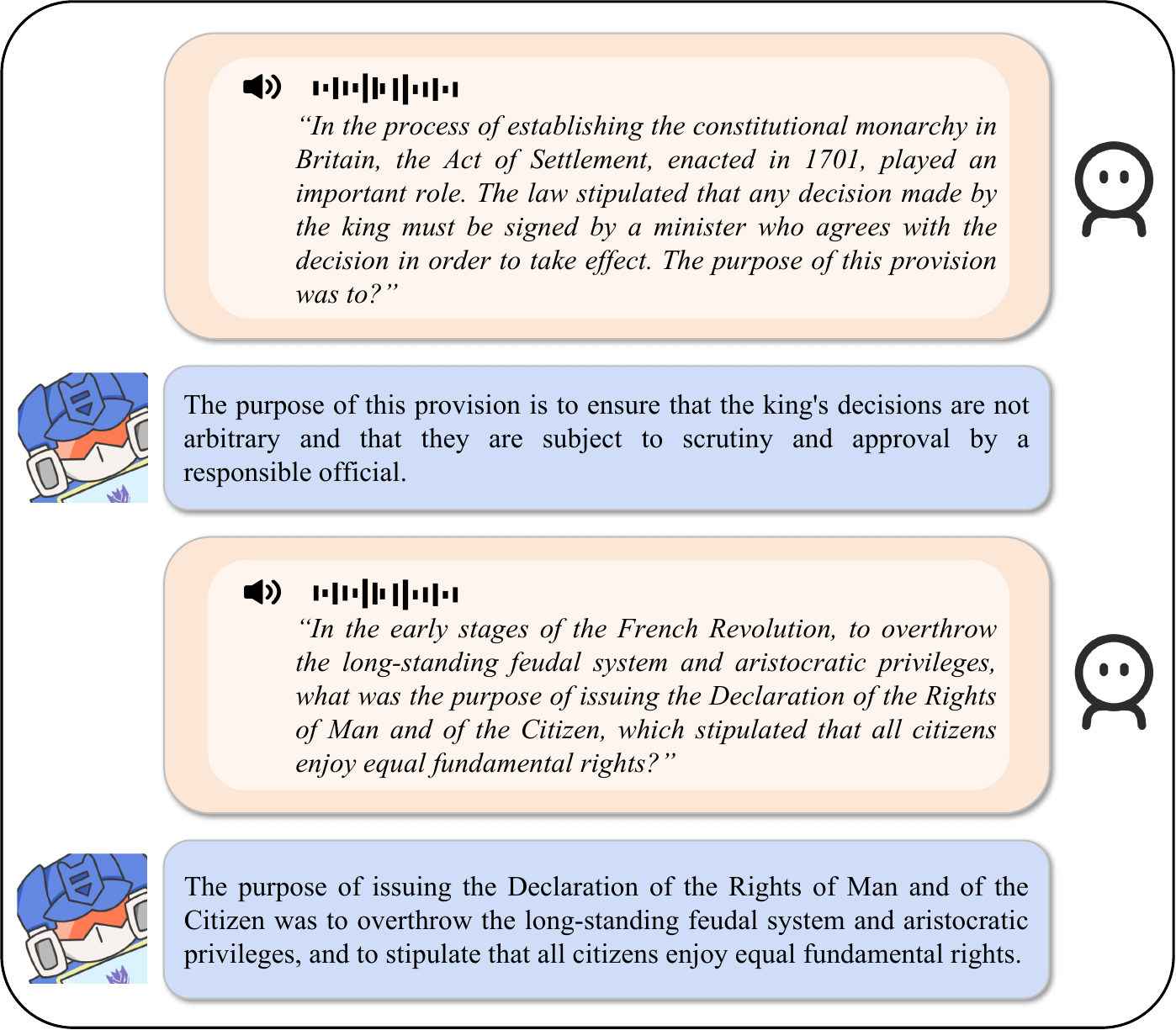}
    \caption{Knowledge-based QA about history.}
    \label{fig: history_case}
\end{figure*}

We will categorize the knowledge to showcase our model's QA ability across various domains of knowledge. Specifically, for math, please refer to Fig.~\ref{fig: math_case}. For physics, please refer to Fig.~\ref{fig: physics_case}. For biology, please refer to Fig.~\ref{fig: biology_case}. For history, please refer to Fig.~\ref{fig: history_case}.

\end{document}

%% file: arch.tex
\begin{figure*}[t]
    \centering
    \small
    \hspace*{-1.2cm}
    \subfigure[Alignment stage]{
    \begin{minipage}[t]{0.24\linewidth}
    \centering
      \begin{tikzpicture} [scale=0.8]
        \node(ae) at (0,0) [rectangle, draw=black, fill=gray!10, rounded corners=3pt, thick, minimum width=1.8cm,minimum height=1cm,align=center] {Audio\\encoder};
        \node(freeze) at ([xshift=0.8cm,yshift=0.3cm]ae.center) [rectangle, align=center] {\Large{\ding{100}}};
        \node(fb) at ([yshift=-0.3cm]ae.south) [rectangle, align=center,anchor=north] {Speech};
        \node(aa) at ([yshift=0.3cm]ae.north) [rectangle, draw=black, fill=orange!10, rounded corners=3pt, thick, minimum width=1.8cm,minimum height=0.5cm,align=center,anchor=south] {Alignment\\adapter};
        
        \node(f1) at ([yshift=1.0cm]aa.west) [rectangle, draw=black, fill=red!10, rounded corners=2pt, thick, minimum width=0.3cm, minimum height=0.3cm,align=center,anchor=west] {};
        \node(f2) at ([xshift=0.2cm]f1.east) [rectangle, draw=black, fill=red!10, rounded corners=2pt, thick, minimum width=0.3cm, minimum height=0.3cm,align=center,anchor=west] {};
        \node(f3) at ([xshift=0.075cm]f2.east) [rectangle, draw=white,  thick, align=center,anchor=west] {...};
        \node(f4) at ([xshift=0.075cm]f3.east) [rectangle, draw=black, fill=red!10, rounded corners=2pt, thick, minimum width=0.3cm, minimum height=0.3cm,align=center,anchor=west] {};
        \node(t1) at ([yshift=-0.05cm]f1.north) [rectangle, align=center,anchor=south] {Nice};
        \node(t2) at ([yshift=-0.05cm]f2.north) [rectangle, align=center,anchor=south] {to};
        \node(t4) at ([yshift=-0.05cm]f4.north) [rectangle, align=center,anchor=south] {<blank>};
        \node(se) at ([xshift=0.075cm,yshift=-0.2cm]f4.east) [rectangle, align=center,anchor=west] {Shared LLM\\embeddings};
        \node(ctc) at ([yshift=1.0cm]aa.north) [rectangle, rounded corners=3pt, thick, align=center,anchor=south] {CTC loss};

        \draw[->,thick]([yshift=-0.05cm]fb.north)--(ae.south);
        \draw[->,thick](ae.north)--(aa.south);
        \draw[->,thick](aa.north)--([yshift=0.3cm]aa.north);

      \end{tikzpicture}
    \end{minipage}
    }
    \subfigure[Shrinking stage]{
    \begin{minipage}[t]{0.45\linewidth}
    \centering
    \begin{tikzpicture} [scale=0.8]
        \node(ae) at (0,0) [rectangle, draw=black, fill=gray!10, rounded corners=3pt, thick, minimum width=1.8cm,minimum height=1cm,align=center] {Audio\\encoder};
        \node(freeze) at ([xshift=0.8cm,yshift=0.3cm]ae.center) [rectangle, align=center] {\Large{\ding{100}}};
        \node(fb) at ([yshift=-0.3cm]ae.south) [rectangle, align=center,anchor=north] {Speech (optional)};
        \node(aa) at ([yshift=0.3cm]ae.north) [rectangle, draw=black, fill=orange!10, rounded corners=3pt, thick, minimum width=1.8cm,minimum height=0.5cm,align=center,anchor=south] {Alignment\\adapter};
        \node(ctc) at ([yshift=0.6cm]aa.north) [rectangle,align=center,anchor=south] {CTC loss};
        \node(sa) at ([xshift=0.4cm,yshift=-0.05cm]ae.east) [rectangle, draw=black, fill=orange!10, rounded corners=3pt, thick, minimum width=1.8cm,minimum height=0.5cm,align=center,anchor=west] {Shrinking\\adapter};
        \node(llm) at ([yshift=1.6cm]sa.west) [rectangle, draw=black, fill=gray!10, rounded corners=3pt, thick, minimum width=2.7cm,minimum height=1.0cm,align=center,anchor=west] {LLMs};
        \node(lora) at (llm.east) [rectangle, draw=black, fill=orange!10, rounded corners=3pt, thick, minimum width=1.0cm,minimum height=1.0cm,align=center,anchor=west] {LoRA};
        \node(te) at ([xshift=0.1cm]sa.east) [rectangle, draw=black, fill=gray!10, rounded corners=3pt, thick, minimum width=1.8cm,minimum height=0.5cm,align=center,anchor=west] {Text\\Embedding};
        \node(freeze3) at ([xshift=0.8cm,yshift=0.2cm]te.center) [rectangle, align=center] {\Large{\ding{100}}};
        \node(ti) at ([yshift=-0.3cm]te.south) [rectangle, align=center,anchor=north] {Text instruction};
        \node(freeze2) at ([xshift=1.2cm,yshift=0.3cm]llm.center) [rectangle, align=center] {\Large{\ding{100}}};
        \node(loss) at ([xshift=0.5cm, yshift=0.3cm]llm.north) [rectangle, align=center,anchor=south] {Next token prediction};

        \draw[->,thick]([yshift=-0.05cm]fb.north)--(ae.south);
        \draw[->,thick](ae.north)--(aa.south);
        \draw[->,thick](aa.north)--(ctc.south);
        \draw[->,thick](sa.north)--([yshift=0.4cm]sa.north);
        \draw[->,thick](te.north)--([yshift=0.4cm]te.north);
        \draw[->,thick]([yshift=-0.3cm]loss.south)--(loss.south);
        \draw[->,thick]([yshift=-0.1cm]ti.north)--(te.south);

        \draw[->,thick](aa.north)--([yshift=0.3cm]aa.north)--([xshift=0.2cm, yshift=0.3cm]aa.north -| aa.east)--([xshift=0.2cm, yshift=-0.3cm]sa.south -| aa.east)--([yshift=-0.3cm]sa.south)--(sa.south);
      \end{tikzpicture}
    \end{minipage}
    }
    \subfigure[SFT stage]{
    \begin{minipage}[t]{0.20\linewidth}
    \centering
    \begin{tikzpicture} [scale=0.8]
        \node(ae) at (0,0) [rectangle, draw=black, fill=gray!10, rounded corners=3pt, thick, minimum width=1.8cm,minimum height=1cm,align=center] {Audio\\encoder};
        \node(freeze) at ([xshift=0.8cm,yshift=0.3cm]ae.center) [rectangle, align=center] {\Large{\ding{100}}};
        \node(fb) at ([yshift=-0.2cm]ae.south) [rectangle, align=center,anchor=north] {Speech};
        \node(aa) at ([yshift=0.2cm]ae.north) [rectangle, draw=black, fill=gray!10, rounded corners=2pt, thick, minimum width=1.8cm,minimum height=0.5cm,align=center,anchor=south] {Two adapters};
        
        \node(llm) at ([yshift=1.1cm]aa.west) [rectangle, draw=black, fill=gray!10, rounded corners=3pt, thick, minimum width=2.7cm,minimum height=0.9cm,align=center,anchor=west] {LLMs};
        \node(lora) at (llm.east) [rectangle, draw=black, fill=orange!10, rounded corners=3pt, thick, minimum width=0.9cm,minimum height=0.9cm,align=center,anchor=west] {LoRA};
        \node(te) at ([xshift=0.1cm,yshift=0.4cm]ae.east) [rectangle, draw=black, fill=gray!10, rounded corners=3pt, thick, minimum width=1.8cm,minimum height=0.5cm,align=center,anchor=west] {Text\\Embedding};
        \node(freeze3) at ([xshift=0.8cm,yshift=0.2cm]te.center) [rectangle, align=center] {\Large{\ding{100}}};
        \node(ti) at ([yshift=-0.4cm]te.south) [rectangle, align=center,anchor=north] {Text instruction};
        \node(freeze2) at ([xshift=1.2cm,yshift=0.2cm]llm.center) [rectangle, align=center] {\Large{\ding{100}}};
        \node(loss) at ([xshift=0.5cm, yshift=0.2cm]llm.north) [rectangle, align=center,anchor=south] {Next token prediction};
       
        \draw[->,thick]([yshift=-0.05cm]fb.north)--(ae.south);
        \draw[->,thick](ae.north)--(aa.south);
        \draw[->,thick](aa.north)--([yshift=0.2cm]aa.north);
        \draw[->,thick](te.north)--([yshift=0.5cm]te.north);
        \draw[->,thick]([yshift=-0.2cm]loss.south)--(loss.south);
        \draw[->,thick]([yshift=-0.1cm]ti.north)--(te.south);
        
      \end{tikzpicture}
    \end{minipage}
    }
      \caption{Training progress of Soundwave. The gray modules are frozen while the orange modules are updated.}
      \label{architecture}
  \end{figure*}

  

%% file: shrinking.tex
\begin{wrapfigure}[13]{r}{0.5\textwidth}
    \vspace{-0.1cm}
  \centering
    \begin{tikzpicture}[scale=0.92]
    \small{
      \node(c) at (0,0) 
      [rectangle, draw=black, fill={rgb,255:red,204; green,204; blue,255},  thick, minimum width=0.3cm,minimum height=0.3cm,rounded corners=1pt,align=center] {};
      \node(l2) at ([xshift=-0.4cm]c.west) 
      [rectangle, draw=black, fill={rgb,255:red,151; green,154; blue,240},  thick, minimum width=0.3cm,minimum height=0.3cm,rounded corners=1pt,align=center] {};  
      \node(l3) at ([xshift=-0.4cm]l2.west) 
      [rectangle, draw=black, fill={rgb,255:red,204; green,204; blue,248},  thick, minimum width=0.3cm,minimum height=0.3cm,rounded corners=1pt,align=center] {};  
      \node(l4) at ([xshift=-0.4cm]l3.west) 
      [rectangle, draw=black, fill={rgb,255:red,104; green,107; blue,234},  thick, minimum width=0.3cm,minimum height=0.3cm,rounded corners=1pt,align=center] {}; 
      \node(l5) at ([xshift=-0.4cm]l4.west) 
      [rectangle, draw=black, fill={rgb,255:red,224; green,224; blue,255},  thick, minimum width=0.3cm,minimum height=0.3cm,rounded corners=1pt,align=center] {}; 
      \node(r1) at ([xshift=0.4cm]c.east) 
      [rectangle, draw=black, fill={rgb,255:red,255; green,255; blue,255},  thick, minimum width=0.3cm,minimum height=0.3cm,rounded corners=1pt,align=center] {}; 
       \node(r2) at ([xshift=0.4cm]r1.east) 
      [rectangle, draw=black, fill={rgb,255:red,255; green,255; blue,255},  thick, minimum width=0.3cm,minimum height=0.3cm,rounded corners=1pt,align=center] {};
       \node(r3) at ([xshift=0.4cm]r2.east) 
      [rectangle, draw=black, fill={rgb,255:red,223; green,239; blue,223},  thick, minimum width=0.3cm,minimum height=0.3cm,rounded corners=1pt,align=center] {};
       \node(r4) at ([xshift=0.4cm]r3.east) 
      [rectangle, draw=black, fill={rgb,255:red,180; green,218; blue,180},  thick, minimum width=0.3cm,minimum height=0.3cm,rounded corners=1pt,align=center] {};
      \node(r5) at ([xshift=0.4cm]r4.east) 
      [rectangle, draw=black, fill={rgb,255:red,211; green,233; blue,211},  thick, minimum width=0.3cm,minimum height=0.3cm,rounded corners=1pt,align=center] {};
      \node(step1) at ([xshift=-0.1cm, yshift=0.45cm]l5.north) 
      [rectangle,  align=center,anchor=east] {{\color{red!30}\ding{192} Select}};
      \node() at([xshift=0.4cm]r5.east)  [rectangle, align=center] {\Large{...}};
      \node() at([xshift=-0.4cm]l5.west)  [rectangle, align=center] {\Large{...}};
      
      \node(q) at ([xshift=0.1cm,yshift=0.9cm]l5.north) [rectangle, draw=red!30, fill=red!10,  thick, minimum width=1.8cm,minimum height=0.5cm,rounded corners=2pt,align=center, anchor=center] {};
      \node(qt1) at (q.north) [rectangle, align=center, anchor=south] {Speech\\content};
      \node(qt2) at ([xshift=0.1cm,yshift=0.2cm]q.east) [rectangle, align=center, anchor=west] {{\color{blue!40}\ding{193} Query}};
      
      \node(q1) at ([xshift=-0.3cm,yshift=0.9cm]l5.north) 
      [rectangle, draw=black, fill={rgb,255:red,104; green,107; blue,234},  thick, minimum width=0.3cm,minimum height=0.3cm,rounded corners=1pt,align=center] {};
      \node(q2) at ([xshift=0.2cm]q1.east) 
      [rectangle, draw=black, fill=white,  thick, minimum width=0.3cm,minimum height=0.3cm,rounded corners=1pt,align=center,anchor=center] {};
      \node(q3) at ([xshift=0.2cm]q2.east) 
      [rectangle, draw=black, fill={rgb,255:red,180; green,218; blue,180},  thick, minimum width=0.3cm,minimum height=0.3cm,rounded corners=1pt,align=center,anchor=center] {};
      
      \node() at([xshift=-0.2cm]q1.west)  [rectangle, align=center] {\large{...}};
      \node() at([xshift=0.2cm]q3.east)  [rectangle, align=center] {\large{...}};

      \node(o) at ([xshift=-0.1cm,yshift=0.9cm]r5.north) [rectangle, draw=blue!50, fill=blue!10,  thick, minimum width=1.8cm,minimum height=0.5cm,rounded corners=2pt,align=center, anchor=center] {};
      \node(ot1) at (o.north) [rectangle, align=center, anchor=south] {Auxiliary\\information};
      \node(ot2) at ([yshift=0.2cm]o.west) [rectangle, align=center, anchor=east] {{\color{blue!40} \& Gather}};

      \node(o1) at ([xshift=-0.5cm,yshift=0.9cm]r5.north) 
      [rectangle, draw=black, fill={rgb,255:red,104; green,107; blue,234},  thick, minimum width=0.3cm,minimum height=0.3cm,rounded corners=1pt,align=center] {};
      \node(o2) at ([xshift=0.2cm]o1.east) 
      [rectangle, draw=black, fill=white,  thick, minimum width=0.3cm,minimum height=0.3cm,rounded corners=1pt,align=center,anchor=center] {};
      \node(o3) at ([xshift=0.2cm]o2.east) 
      [rectangle, draw=black, fill={rgb,255:red,180; green,218; blue,180},  thick, minimum width=0.3cm,minimum height=0.3cm,rounded corners=1pt,align=center,anchor=center] {};
      
      \node() at([xshift=-0.2cm]o1.west)  [rectangle, align=center] {\large{...}};
      \node() at([xshift=0.2cm]o3.east)  [rectangle, align=center] {\large{...}};

      \node(output) at ([xshift=0.2cm,yshift=1.7cm]c.north) [rectangle, align=center, anchor=south] {{\color{red!30}Out}{\color{blue!40}put}};

      \draw[decorate, decoration={brace,amplitude=2mm}, thick] ([yshift=-0.15cm]c.south) -- ([yshift=-0.15cm]l5.south);
      
      \draw[decorate, decoration={brace,amplitude=2mm}, thick] ([xshift=0.2cm, yshift=-0.15cm]r2.south) -- ([xshift=-0.2cm, yshift=-0.15cm]r1.south);

      \draw[decorate, decoration={brace,amplitude=2mm}, thick] ([xshift=0.2cm, yshift=-0.15cm]r5.south) -- ([xshift=-0.2cm, yshift=-0.15cm]r3.south);
      
      \node(nice) at ([yshift=-0.55cm]l3.south) [rectangle, align=center] {Nice};

      \node(blank) at ([xshift=0.3cm, yshift=-0.55cm]r1.south) [rectangle, align=center] {<blank>};
      \node(to) at ([yshift=-0.55cm]r4.south) [rectangle, align=center] {to};
      
      \draw[->,thick,color=red!30](l4.north).. controls ([yshift=0.3cm]l4.north) and ([yshift=-0.3cm]q1.south) ..(q1.south);
      \draw[->,thick,color=red!30](r1.north).. controls ([xshift=-0.2cm, yshift=0.4cm]r1.north) and ([xshift=0.2cm,yshift=-0.4cm]q2.south) ..(q2.south);
      \draw[->,thick,color=red!30](r4.north).. controls ([yshift=0.55cm]r4.north) and ([xshift=0.1cm,yshift=-0.5cm]q3.south) ..([xshift=0.1cm]q3.south);
      
      \draw[->,thick,color=blue!40](q.east).. controls ([yshift=-0.3cm]qt2.center)..([xshift=0.6cm,yshift=-0.55cm]qt2.south);
       \node() at ([xshift=0.3cm,yshift=0.05cm]c) [rectangle, draw=blue!30, thick, minimum width=6.4cm,minimum height=0.7cm,rounded corners=2pt,align=center, anchor=center] {};
       \draw[->,thick,color=blue!40] ([xshift=0.6cm,yshift=-0.55cm]qt2.south).. controls ([xshift=1.0cm,yshift=-0.3cm]qt2.center).. (o.west);

       \draw[->,thick,color=red!30]([yshift=0.25cm,xshift=-0.1cm]q.east).. controls ([xshift=0.2cm,yshift=0.7cm]q.east) and ([xshift=-0.2cm,yshift=-0.5cm]output.south) ..([yshift=0.1cm]output.south);
       \draw[->,thick,color=blue!40]([yshift=0.25cm,xshift=0.1cm]o.west).. controls ([xshift=-0.2cm,yshift=0.7cm]o.west) and ([xshift=0.2cm,yshift=-0.5cm]output.south) ..([yshift=0.1cm]output.south);
      }
    \end{tikzpicture}

    \caption{We first select the features based on the peak of CTC prediction. Then, we use these features to query and gather auxiliary information from the original sequence. Finally, we fuse the two features to achieve shrinking. }
    \label{lbm_method}
    \vspace{-0.2cm}
\end{wrapfigure}

%% file: data_info.tex
\begin{table*}[th]
\centering
\small
\setlength{\tabcolsep}{5pt} 
\caption{Summary of datasets used in different stages and their total hours.}
\label{datasets}
\resizebox{1.\textwidth}{!}{
\begin{threeparttable}
\begin{tabular}{l c c c r r l}
\toprule
\textbf{Dataset} & \textbf{I} & \textbf{II} & \textbf{III} & \textbf{Num.} & \textbf{Hours} & \multicolumn{1}{c}{\textbf{Task}} \\
\midrule
GigaSpeech (M) \citep{gigaSpeech} & \checkmark & \checkmark &  & 713k & 805.11 & ASR\\
TED-LIUM \citep{TED-LIUM} & \checkmark & \checkmark &  & 144k & 244.02 & ASR\\
Multilingual Librispeech (En) \citep{MLS} & \checkmark & \checkmark &  & 985k & 4,081.61 & ASR\\
Europarl-ASR \citep{europarlASR} & \checkmark & \checkmark &  & 719k & 418.42 & ASR\\
TextrolSpeech \citep{TextrolSpeech} & \checkmark & \checkmark & \checkmark & 215k & 301.19 & ASR, GR~\tnote{1}, Emotion Recognition \\
LibriSpeech \citep{LibriSpeech} & \checkmark & \checkmark & \checkmark & 281k & 961.05 & ASR, Speech Grounding \\
MUST-C (En-De) \citep{MuST-C} & & \checkmark & \checkmark & 283k & 388.55 & Speech Translation\\
Common Voice (En) \citep{ardila2019common_voice} & & \checkmark & \checkmark & 233k & 364.64 & AP~\tnote{1}, Speech Translation \\
Fisher \citep{cieri2004fisher} & & \checkmark & \checkmark & 132k & 1,091.42 & ASR, Chat \\
Europarl-ST \citep{Europarl-ST} & & & \checkmark & 53k & 133.16 & Language Identification \\
Common Voice (Ja) \citep{ardila2019common_voice} & & & \checkmark & 13k & 15.00 & Language Identification \\
SLURP \citep{SLURP} & & & \checkmark & 141k & 101.49 & IC~\tnote{1}, Entity Recognition \\
CREMA-D \citep{Crema-d} & & & \checkmark & 7k & 5.26 & Emotion Recognition \\
RAVDESS \citep{RAVDESS} & & & \checkmark & 1k & 1.48 & Emotion Recognition \\
IEMOCAP \citep{IEMOCAP} & & & \checkmark & 3k & 2.16 & Emotion Recognition \\
MELD \citep{MELD} & & & \checkmark & 9k & 8.12 & Emotion Recognition \\
VoxCeleb \citep{VoxCeleb} & & & \checkmark & 156k & 435.17 & Speaker Num. Verification \\
FoR \citep{FoR} & & & \checkmark & 54k & 47.55 & Synthesized Detection \\
AnyInstruct \citep{zhan2024anygpt} & & & \checkmark & 107k & 206.30 & Speech Instruction \\
\hline
VocalSound \citep{VocalSound} & \checkmark & \checkmark & \checkmark & 20k & 23.20 & Sound Classification \\
TUT2017 \citep{TUTscene} & \checkmark & \checkmark & \checkmark & 5k & 13.00 & Scene Classification \\
CochlScene \citep{CochlScene} & & & \checkmark & 75k & 208.65 & Scene Classification \\
\midrule
\textbf{Total}~\tnote{2} & & & & \textbf{4,349k$^*$}\textsuperscript{} & \textbf{9,856.91$^*$}\textsuperscript{} & \textbf{15} \\
\bottomrule
\end{tabular}

\end{threeparttable}
}
\begin{tablenotes}[para,flushleft]
\item [1]  GR is for Gender Recognition, AP is for Age Prediction, and IC is for Intent Classification. \\
\item [2] `*' means that this table is compiled from the perspective of audio, and an audio file may be used multiple times\\for different tasks. If multiple usages at different tasks are all counted, the number of data samples is 6301k, and \\the total duration is 14068.77 hours.
\end{tablenotes}
\end{table*}

%% file: model_compare_loss.tex
\begin{wrapfigure}[9]{r}{0.47\textwidth}
    \centering
    \vspace{-0.3cm}
    \begin{tikzpicture}
        \pgfplotsset{set layers}
        \scriptsize{
        \begin{axis}[at={(0,0)},
            ymajorgrids,
            xmajorgrids,
            grid style=dashed,
            width=.5\textwidth,
            height=.25\textwidth,
            legend style={at={(0.04,0.92)},legend image code/.code={
            \draw[#1] (0cm,0cm) -- (0.2cm,0cm);}, anchor=south west},
            xlabel={\scriptsize{Training steps}},
            ylabel={\scriptsize{Loss}},
            ylabel style={xshift=0em,yshift=-0.5em},
            yticklabel style={/pgf/number format/precision=3, /pgf/number format/fixed},
            ymin=0, ymax=20, ytick={5,10,15},
            xmin=10, xmax=210, xtick={20,60,100,140,180},
            legend columns=2,
            legend style={draw=gray!50, yshift=-2.9em, xshift=4.8em, cells={anchor=west}, fill opacity=0.8}
        ]
          
        \addplot[nasdaqup!70, mark=o*, mark size=1.0pt, thick, mark options={fill=white, draw=nasdaqup, line width=0.5pt}] coordinates {(20, 2.8256) (25, 2.3121) (30, 1.7106) (35, 1.2081) (40, 0.9505) (45, 0.8636) (50, 0.7716) (55, 0.7397) (60, 0.7039) (65, 0.6605) (70, 0.6821) (75, 0.6283) (80, 0.6153) (85, 0.5804) (90, 0.5847) (95, 0.5703) (100, 0.5613) (105, 0.5719) (110, 0.531) (115, 0.5193) (120, 0.4951) (125, 0.4928) (130, 0.497) (135, 0.4808) (140, 0.4693) (145, 0.452) (150, 0.453) (155, 0.4303) (160, 0.4376) (165, 0.4411) (170, 0.4184) (175, 0.4305) (180, 0.4219) (185, 0.4169) (190, 0.4333) (195, 0.4211) (200, 0.4393)}; 
        \addlegendentry{\scalebox{.8}{Soundwave}}
        
        \addplot[dodgerblue!70, mark=o*, mark size=1.0pt, thick, mark options={fill=white, draw=dodgerblue, line width=0.5pt}] coordinates {(20, 13.0869) (25, 18.2169) (30, 9.5901) (35, 8.7578) (40, 8.351) (45, 9.4212) (50, 7.5364) (55, 7.3697) (60, 7.0582) (65, 6.6255) (70, 6.4926) (75, 6.2211) (80, 6.0975) (85, 5.9839) (90, 5.8997) (95, 6.4775) (100, 6.163) (105, 6.3109) (110, 5.7242) (115, 5.7483) (120, 5.6953) (125, 5.6869) (130, 5.6733) (135, 6.551) (140, 6.0559) (145, 5.9298) (150, 5.6268) (155, 5.6167) (160, 5.5582) (165, 5.485) (170, 5.7994) (175, 9.2947) (180, 5.4763) (185, 5.7133) (190, 5.9157) (195, 5.5137) (200, 5.495)};
        \addlegendentry{\scalebox{.8}{Soundwave \textit{w/o} Stage I}}

        \addplot[
            orangered!70,
            mark=o*,
            mark size=1.0pt,
            thick,
            mark options={fill=white, draw=orangered, line width=0.5pt}
        ] coordinates {
            (20, 4.0585) (25, 3.7696) (30, 3.55) (35, 3.383) (40, 3.2954) (45, 3.2508) (50, 3.2044) (55, 3.1427) (60, 3.1292) (65, 3.089) (70, 3.0967) (75, 3.076) (80, 3.0665) (85, 3.053) (90, 3.0388) (95, 3.0374) (100, 3.0405) (105, 3.028) (110, 3.0239) (115, 3.0134) (120, 2.9823) (125, 3.0036) (130, 2.957) (135, 2.9561) (140, 2.9444) (145, 2.9355) (150, 2.9231) (155, 2.9227) (160, 2.9011) (165, 2.8939) (170, 2.868) (175, 2.8679) (180, 2.8614) (185, 2.8544) (190, 2.8526) (195, 2.8437) (200, 2.8259)
        };
        \addlegendentry{\scalebox{.8}{Adapter ($\times3$)}}
        \end{axis}}
    \end{tikzpicture}
    \caption{Training curves of different strategies}
    \label{convergence_speed}
    \vspace{-0.3cm}
\end{wrapfigure}

%% file: compare_sim_and_train_speed.tex
\begin{wrapfigure} [11]{r}{0.47\textwidth}
  \centering
  \vspace{-0.4cm}
  \begin{tikzpicture}
    \scriptsize
    \begin{axis}[
      ymajorgrids,
      grid style=dashed,
      ybar=3pt,
      enlarge x limits=0.5,
      xtick align=inside,
      height=.25\textwidth,
      width=.27\textwidth, 
      bar width=1.2em,
      xlabel={(a) Feature similarity between\\ audio and text},
      xlabel style={xshift=-0.3cm},
      ylabel={Similarity ($10^{-2}$)},
      xtick=data,
      symbolic x coords={A-T}, 
      nodes near coords,  
      nodes near coords align={vertical},
      ymin=0,
      ymax=60,
      ytick={10,30,50},
      xticklabels={
        {} 
      },
      x tick label style={align=center, font=\tiny}, 
      xlabel style={yshift=0.3em,align=center},
      yticklabel style={,rotate=90},
    ]
    every node near coord/.append style={  
        font=\tiny,  
        /pgf/number format/.cd,  
        fixed,  
        fixed zerofill,  
        precision=2  
    }

      \addplot[bar shift=-2em, fill=red!30, draw=red, area legend] coordinates {
        (A-T, 48.8) 
      };
      \addplot[fill=teal!30, draw=teal, area legend] coordinates {
        (A-T, 3.8) 
      };
      \addplot[bar shift=2em, fill=blue!30, draw=blue, area legend] coordinates {
        (A-T, 3.6) 
      };
      \legend{};
    \end{axis}
    
    \scriptsize{
    \begin{axis}[
      at={(13.5em,0)},
      ymajorgrids,
      grid style=dashed,
      legend style={draw=gray!70, at={(-1.535,1.03)}, anchor=south west},
      legend columns=3,
      legend cell align={left},
      ybar,
      enlarge x limits=0.5,
      xtick align=inside,
      height=.25\textwidth,
      width=.27\textwidth,
      bar width=1.2em,
      xlabel={(b) \ Training speed on different\\compression method},
      ylabel={s / step},
      xlabel style={xshift=-0.2cm},
      symbolic x coords={{1}, {2}},
      xtick=data,
      nodes near coords,
      nodes near coords align={vertical},
      ymin=0,
      ymax=104,
      ytick={20, 40, 60, 80},
      xticklabels={},
      legend entries={Soundwave, Adapter ($\times3$), Adapter ($\times4$)},
      xlabel style={yshift=0.3em,align=center},
      yticklabel style={rotate=90},
      ]
      \addplot[bar shift=-2em, fill=red!30, draw=red, area legend] coordinates {
        ({1}, 25.8) 
      };
      \addlegendentry{\scalebox{.8}{Soundwave}}

      \addplot[fill=teal!30, draw=teal,area legend] coordinates {({1},86.7)};
      \addlegendentry{\scalebox{.8}{Adapter ($\times3$)}} 
      
      \addplot[bar shift=2em, fill=blue!30, draw=blue,area legend] coordinates {({1},72.4)}; 
      \addlegendentry{\scalebox{.8}{Adapter ($\times4$)}}  
    \end{axis}
    }
\end{tikzpicture}
\vspace{-0.5cm}
\caption{Comparison of alignment effects and training speeds of different methods.}
\label{denosieperformance}

\end{wrapfigure} 

%% file: data_quality_loss_compare.tex
\begin{wrapfigure} [10]{r}{0.48\textwidth}
    \centering
    \vspace{-0.4cm}
    \begin{tikzpicture}
        \pgfplotsset{set layers}
        \scriptsize{
        \begin{axis}[at={(0,0)},
            ymajorgrids,
            xmajorgrids,
            grid style=dashed,
            width=.5\textwidth,
            height=.25\textwidth,
            legend style={at={(0.04,0.90)}, legend image code/.code={
            \draw[#1] (0cm,0cm) -- (0.2cm,0cm);}, anchor=south west},
            xlabel={\scriptsize{Training steps}},
            ylabel={\scriptsize{Loss}},
            ylabel style={xshift=0em,yshift=-0.5em},
            yticklabel style={/pgf/number format/precision=3, /pgf/number format/fixed},
            ymin=0, ymax=25, ytick={5,10,15,20},
            xmin=50, xmax=500, xtick={100,200,300,400,500},
            legend columns=2,
            legend style={draw=gray!50, yshift=-18.0pt, xshift=2.8em, cells={anchor=west}, fill opacity=0.8}
        ]
          
        \addplot[nasdaqup!70, mark=o*, mark size=1.0pt, thick, mark options={fill=white, draw=nasdaqup, line width=0.5pt}] coordinates {(50, 10.5829) (60, 9.9488) (70, 8.9557) (80, 13.9119) (90, 8.9847) (100, 9.3311) (110, 9.543) (120, 8.4029) (130, 9.3018) (140, 8.0244) (150, 7.9054) (160, 9.6434) (170, 8.4956) (180, 8.5991) (190, 8.0765) (200, 7.8302) (210, 9.1939) (220, 7.7429) (230, 7.7262) (240, 7.6379) (250, 8.1192) (260, 8.0659) (270, 7.7758) (280, 7.8278) (290, 7.5231) (300, 7.5616) (310, 7.425) (320, 7.4722) (330, 7.4812) (340, 7.3787) (350, 7.3357) (360, 7.3153) (370, 7.2285) (380, 7.1661) (390, 7.503) (400, 7.7288) (410, 6.8461) (420, 6.8559) (430, 6.319) (440, 6.9405) (450, 6.2354) (460, 5.4715) (470, 4.9763) (480, 4.7738) (490, 4.5941) (500, 4.065)}; 
        \addlegendentry{\scalebox{.8}{Speech + Sound}}
        
        \addplot[dodgerblue!70, mark=o*, mark size=1.0pt, thick, mark options={fill=white, draw=dodgerblue, line width=0.5pt}] coordinates {(50, 12.6348) (60, 14.3148) (70, 9.4531) (80, 9.8651) (90, 10.3501) (100, 9.8556) (110, 13.8207) (120, 15.3878) (130, 10.7466) (140, 10.9202) (150, 9.5874) (160, 9.4762) (170, 10.0022) (180, 17.6529) (190, 14.0038) (200, 10.0795) (210, 9.456) (220, 9.4732) (230, 9.7444) (240, 8.6563) (250, 10.6683) (260, 8.3631) (270, 8.3816) (280, 8.3815) (290, 8.2367) (300, 9.2573) (310, 8.4542) (320, 10.2458) (330, 8.9888) (340, 8.6106) (350, 9.0814) (360, 9.2304) (370, 8.5569) (380, 8.5301) (390, 8.1311) (400, 8.0919) (410, 8.4245) (420, 8.4108) (430, 8.2323) (440, 8.0144) (450, 8.1056) (460, 9.511) (470, 8.3807) (480, 7.8865) (490, 10.9731) (500, 8.3905)};
        \addlegendentry{\scalebox{.8}{Speech + Uncleaned sound}}

        \addplot[
            orangered!70,
            mark=o*,
            mark size=1.0pt,
            thick,
            mark options={fill=white, draw=orangered, line width=0.5pt}
        ] coordinates {
            (50, 13.5313) (60, 11.0681) (70, 11.688) (80, 9.8605) (90, 9.1852) (100, 9.2918) (110, 9.5979) (120, 9.5102) (130, 9.4046) (140, 8.79) (150, 8.9539) (160, 8.4985) (170, 8.5089) (180, 8.6079) (190, 8.3891) (200, 8.4096) (210, 8.2361) (220, 8.096) (230, 7.8528) (240, 7.6303) (250, 7.3708) (260, 6.9564) (270, 6.8045) (280, 6.4365) (290, 5.9436) (300, 5.9387) (310, 5.4209) (320, 5.302) (330, 4.9502) (340, 4.8534) (350, 4.6972) (360, 4.4211) (370, 5.0054) (380, 4.2669) (390, 4.0685) (400, 4.0507) (410, 3.9783) (420, 4.0061) (430, 3.7421) (440, 3.6257) (450, 3.4565) (460, 3.54) (470, 3.3021) (480, 3.4029) (490, 3.4371) (500, 3.6803)
        };
        \addlegendentry{\scalebox{.8}{Speech}}
        \addplot[
            blue!70,
            mark=o*,
            mark size=1.0pt,
            thick,
            mark options={fill=white, draw=orangered, line width=0.5pt}
        ] coordinates {
            (50, 12.1776) (60, 12.1041) (70, 10.7928) (80, 9.6658) (90, 10.3362) (100, 9.8189) (110, 9.8886) (120, 9.1259) (130, 9.5937) (140, 9.8142) (150, 9.4765) (160, 10.3601) (170, 9.9892) (180, 9.6522) (190, 9.1685) (200, 8.9826) (210, 8.4508) (220, 8.3449) (230, 8.6478) (240, 8.2408) (250, 8.2063) (260, 8.3904) (270, 8.9355) (280, 8.281) (290, 9.1995) (300, 8.8117) (310, 14.8669) (320, 10.1081) (330, 20.0605) (340, 109.5972) (350, 21.4714) (360, 9.9985) (370, 17.2364) (380, 27.4544) (390, 14.3754) (400, 9.3341) (410, 7.9659) (420, 9.5716) (430, 7.7453) (440, 8.4507) (450, 7.3517) (460, 7.2264) (470, 8.7026) (480, 6.6073) (490, 7.8524) (500, 5.841)
        };
        \addlegendentry{\scalebox{.8}{Uncleaned speech}}
        \end{axis}}
    \end{tikzpicture}
    \vspace{-0.2cm}
    \caption{Training curves with cleaned and  uncleaned data at the alignment stage.}
    \label{data_quality_convergence_speed}
\end{wrapfigure}

%% file: model_compare_air_speech.tex
\begin{wrapfigure}[10]{r}{0.48\textwidth}
    
\vspace{-0.2cm}
    \centering
    \begin{tikzpicture}
        \pgfplotsset{set layers}
        \scriptsize{
        \begin{axis}[at={(0,0)},
            ymajorgrids,
            xmajorgrids,
            grid style=dashed,
            width=.50\textwidth,
            height=.255\textwidth,
            legend style={at={(-0.095,0.87)}, anchor=south west},
            ylabel={\scriptsize{Acc. (\%)}},
            ylabel style={xshift=0em,yshift=-0.5em},
            yticklabel style={/pgf/number format/precision=3, /pgf/number format/fixed},
            ymin=10, ymax=115, ytick={30,50,70,90},
            xmin=0.5, xmax=9.5, xtick={1,2,3,4,5,6,7,8,9},
            xticklabels={
                SG,
                SLI,
                SGR,
                ER,
                SAP,
                SER,
                IC,
                SNV,
                SVD
            },
            xticklabel style={align=center, font=\tiny, rotate=0},
            legend columns=3,
            legend style={draw=gray!50, yshift=10pt, xshift=0.50em, cells={anchor=west}, fill opacity=0.8}
        ]
          
        \addplot[nasdaqup!70, mark=*, mark size=0.7pt, thick, mark options={fill=nasdaqup, draw=nasdaqup, line width=0.5pt}] coordinates {(1, 59.2) (2, 89.6) (3, 90.3) (4, 60.5) (5, 58.9) (6, 81.7) (7, 93.2) (8, 73.3) (9, 72.5)}; 
        \addlegendentry{\scalebox{.8}{Soundwave (10k)}}
        
        \addplot[dodgerblue!70, mark=*, mark size=0.7pt, thick, mark options={fill=dodgerblue, draw=dodgerblue, line width=0.5pt}] coordinates {(1, 25.2) (2, 32.6) (3, 66.8) (4, 24.6) (5, 50) (6, 33.1) (7, 35.8) (8, 24.6) (9, 51)};
        \addlegendentry{\scalebox{.8}{Soundwave (1k)}}

        \addplot[
            orangered!70,
            mark=*,
            mark size=0.7pt,
            thick,
            mark options={fill=orangered, draw=orangered, line width=0.5pt}
        ] coordinates {
            (1, 25.3) (2, 28.1) (3, 35.5) (4, 29.9) (5, 48.7) (6, 51.7) (7, 36.7) (8, 34.3) (9, 50)
        };
        \addlegendentry{\scalebox{.8}{SALMONN}}
        \end{axis}}
    \end{tikzpicture}
    \caption{Comparison of scaling effect in AIR-Bench speech foundation tasks.}
    \label{scaling_compare}
\end{wrapfigure}